%% file: main.tex
\documentclass[sigconf]{acmart}
\AtBeginDocument{%
  }

\usepackage{amsthm}
\usepackage{amsmath}
\usepackage{subcaption}
\usepackage{paralist}
\usepackage{cases}
\usepackage{booktabs}
\usepackage{threeparttable}
\usepackage{epstopdf}
\usepackage{upgreek}
\usepackage{endnotes}
\usepackage{etoolbox}
\usepackage{algpseudocode}
\usepackage{xspace}
\usepackage{array}
\usepackage{enumitem}
\usepackage{balance}
\usepackage{multirow}
\usepackage{xcolor}
\usepackage{graphicx}
\usepackage{wrapfig}
\usepackage{enumitem}
\usepackage{tabularx}
\usepackage{float}
\usepackage{svg}
\usepackage{subcaption}
\usepackage[ruled,linesnumbered]{algorithm2e}
\begin{document}
\title{An LLM-Powered Cooperative Framework for Large-Scale Multi-Vehicle Navigation}

\author{Yuping Zhou}
\affiliation{
\country{}
\institution{The Hong Kong University of Science and Technology (Guangzhou)}}
\email{yzhou169@connect.hkust-gz.edu.cn}

\author{Siqi Lai}
\affiliation{
\country{}
\institution{The Hong Kong University of Science and Technology (Guangzhou)}}
\email{slai125@connect.hkust-gz.edu.cn}

\author{Jindong Han}
\affiliation{
\country{}
\institution{The Hong Kong University of Science and Technology (Guangzhou)}}
\email{jhanao@connect.ust.hk}

\author{Hao Liu}
\affiliation{
\country{}
\institution{The Hong Kong University of Science and Technology (Guangzhou)}
}
\email{liuh@ust.hk}

\renewcommand{\shortauthors}{Yuping Zhou, Siqi Lai, Jindong Han, & Hao Liu}

\renewcommand{\shortauthors}{Yuping et al.}

\newcommand{\fix}{\marginpar{FIX}}
\newcommand{\new}{\marginpar{NEW}}
\newcommand{\ie}{\emph{i.e.,}\xspace}
\newcommand{\eg}{\emph{e.g.,}\xspace}
\newcommand{\etc}{\emph{etc.}\xspace}
\newcommand{\etal}{\emph{et al.}\xspace}
\newcommand{\TODO}[1]{{\color{red}TODO: {#1}}}
\newcommand{\red}[1]{{\color{red}{#1}}}
\newcommand{\hao}[1]{{\color{blue}{#1}}}
\newcommand{\siqi}[1]{{\color{purple}{#1}}}
\newcommand{\han}[1]{{\color{blue}{#1}}}
\newcommand\beftext[1]{{\color[rgb]{0.5,0.5,0.5}{BEFORE:#1}}}
\newcommand{\eat}[1]{}

\begin{abstract}
The rise of Internet of Vehicles (IoV) technologies is transforming traffic management from isolated control to a collective, multi-vehicle process. At the heart of this shift is multi-vehicle dynamic navigation, which requires simultaneously routing large fleets under evolving traffic conditions. Existing path search algorithms and reinforcement learning methods struggle to scale to city-wide networks, often failing to capture the nonlinear, stochastic, and coupled dynamics of urban traffic.
To address these challenges, we propose \textit{CityNav}, a hierarchical, LLM-powered framework for large-scale multi-vehicle navigation. CityNav integrates a global traffic allocation agent, which coordinates strategic traffic flow distribution across regions, with local navigation agents that generate locally adaptive routes aligned with global directives. To enable effective cooperation, we introduce a \textit{cooperative reasoning optimization} mechanism, in which agents are jointly trained with a dual-reward structure: individual rewards promote per-vehicle efficiency, while shared rewards encourage network-wide coordination and congestion reduction.
Extensive experiments on four real-world road networks of varying scales (up to 1.6 million roads and 430,000 intersections) and traffic datasets demonstrate that CityNav consistently outperforms nine classical path search and RL-based baselines in city-scale travel efficiency and congestion mitigation. Our results highlight the potential of LLMs to enable scalable, adaptive, and cooperative city-wide traffic navigation, providing a foundation for intelligent, large-scale vehicle routing in complex urban environments. Our project is available at \url{https://github.com/usail-hkust/CityNav}.
\end{abstract}  

\ccsdesc[500]{Computing methodologies~Multi-agent planning}

\keywords{internet of vehicles, vehicle navigation, large language model, multi-agent control}


\maketitle

\input{sections/intro}
\input{sections/preliminary}
\input{sections/method}
\input{sections/exp}
\input{sections/related_work}
\input{sections/conclusion}

\bibliographystyle{ACM-Reference-Format}
\bibliography{citations}

\input{sections/appendix}

\end{document}

%% file: sections/intro.tex
\section{Introduction}

The rapid development of Internet of Vehicles (IoV)~\cite{wang2019survey, ji2020survey} technologies is transforming traffic control from an isolated activity into a collective process involving large fleets of connected vehicles. At the heart of this transition lies the problem of multi-vehicle dynamic navigation (MDN)~\cite{koh2020real, wang2022xrouting, yin2025cooperative}, which aims to simultaneously route large fleets of vehicles under continuously evolving traffic conditions. Unlike single-vehicle navigation, MDN is a large-scale distributed coordination task, where each routing decision not only affects an individual vehicle but also alters the overall traffic flow across the network. This shift redefines the goal of navigation, moving beyond individual mobility optimization toward enhancing network-wide traffic efficiency. Despite recent advances, current approaches often fail to meet the demand of scalability, responsiveness, and cooperative decision-making, which are critical for real-world, city-scale deployment.

\begin{figure}[t]
    \centering
    \includegraphics[width=\linewidth]{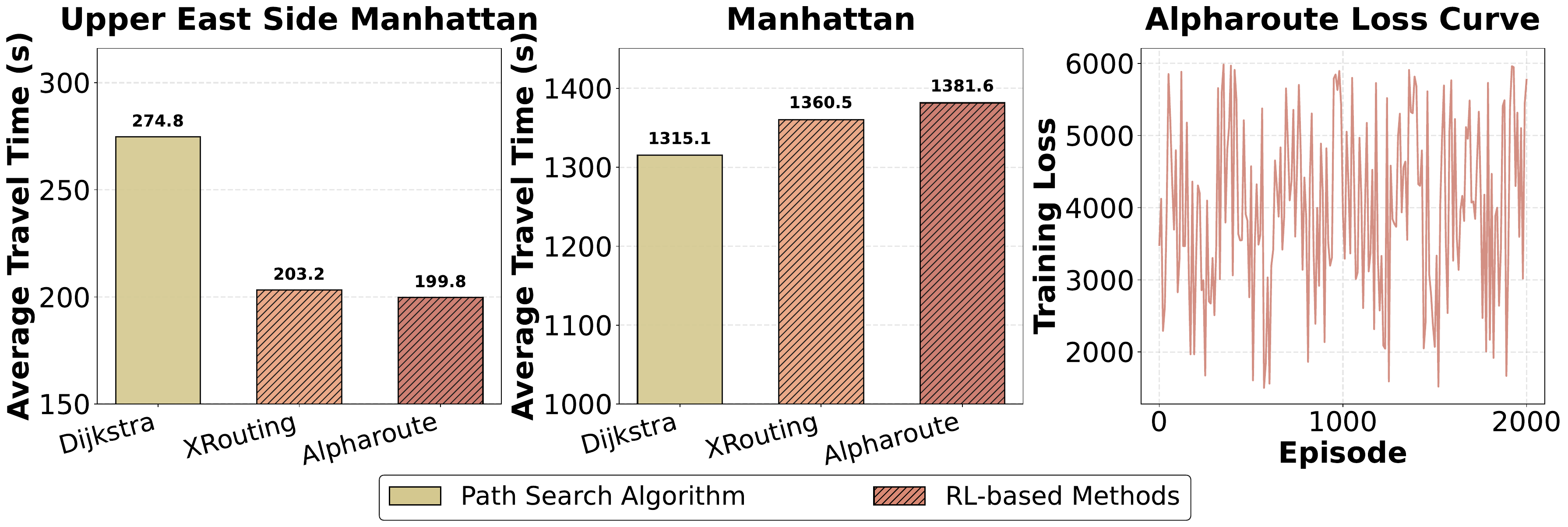}
    \caption{RL performance across different network scales. The Upper East Side covers about one-seventh of Manhattan.}
    \label{fig:Scalability}
\vspace{-10pt}
\end{figure}

Existing approaches to vehicle navigation mainly fall into two paradigms. (1) \emph{Shortest-path search methods} rely on classical algorithms such as Dijkstra~\cite{dijkstra2022note} and A*~\cite{hart1968formal}, which compute optimal routes based on static or simplified cost functions. Their multi-vehicle extensions~\cite{ijcai2021p503, https://doi.org/10.1111/mice.12577, LECLERCQ2021118} attempt to coordinate fleets by injecting congestion penalties or capacity constraints. However, such methods remain largely rule-driven and depend on fixed assumptions about traffic dynamics. Thus, they struggle to handle the nonlinear, stochastic, and time-varying nature of real-world urban traffic~\cite{WANG2022103444}, often producing brittle or inefficient routing outcomes. (2) \emph{Reinforcement learning (RL) methods} model navigation as a sequential decision-making process, where models learn routing policies through interactions with simulated traffic environments~\cite{wang2022xrouting, yin2025cooperative, koh2020real, sun2023hierarchical}. Although RL offers the advantage of adaptive learning and dynamic feedback, it suffers from fundamental scalability bottlenecks. As the number of vehicles or the size of the road network grows, the joint state-action space expands combinatorially, making efficient policy exploration and stable convergence extremely difficult. As shown in Figure \ref{fig:Scalability}, RL methods often perform competitively on small networks but experience sharp performance degradation at the city scale, sometimes even being outperformed by simple shortest-path baselines.

To overcome these limitations, we explore the potential of large language models (LLMs), which have recently demonstrated strong reasoning, flexible communication, and generalization capabilities across diverse domains~\cite{su2024many, huang2025biomni, liu2025mm}. Unlike traditional routing algorithms that rely on hand-crafted heuristics or RL methods that depend on costly trial-and-error training, LLMs provide a new foundation for adaptive, communication-driven coordination. With their vast pretrained knowledge and natural language reasoning capability, LLMs can integrate heterogeneous information, interpret evolving traffic conditions, and generate context-aware routing strategies. More importantly, LLMs inherently possess the ability to communicate and collaborate through language, allowing them to share information, negotiate priorities, and align local decisions with global objectives.
Consequently, LLMs show great promise to overcome the scalability bottlenecks and coordination inefficiencies that constrain existing approaches.

Recent studies have begun to explore LLMs for multi-agent systems, showing that collaborative agents can decompose and solve complex tasks through structured communication. For example, MetaGPT~\cite{hong2024metagpt} deploys four small-scale LLM agents organized into specialized roles to automate software development. In traffic management, CoLLMLight~\cite{yuan2025collmlight} employs up to five hundred agents leveraging spatiotemporal graphs for traffic signal coordination, while CoMAL~\cite{yao2025comal} investigates small-scale collaborative traffic control scenarios with light traffic volumes. These works underscore the potential of LLM-driven coordination. However, unlike small-scale  coordination, large-scale dynamic navigation involves continuous decision-making across tons of interacting vehicles, where both the \emph{dimensionality of real-time data} and the \emph{coupling between local and global dynamics} pose formidable challenges. Specifically, two critical questions emerge:
(1) How can LLM agents exchange information and generate cooperative routing strategies while remain computationally feasible at city scale?
(2) How can we align each agent’s local routing decisions with the network-wide objective of minimizing congestion and maximizing overall mobility?

To tackle the above challenges, we propose CityNav, a scalable cooperative framework for network-wide dynamic navigation powered by LLM agents. CityNav adopts an efficient hierarchical architecture that integrates global coordination with local adaptability. At the macro level, a global traffic allocation agent aggregates network-wide information (\eg congestion patterns and demand distributions), and formulates high-level directives to balance traffic loads across partitioned regions. At the micro level, a network of regional navigation agents manages routing within their assigned regions, generating intra-region paths that follow the global agent’s general directives while adapting to real-time local conditions. This division of responsibilities dramatically reduces complexity. The global agent focuses on strategic, network-wide decisions, while local agents handle smaller, localized navigation problems. By narrowing each agent’s decision space and enabling multi-level communication, CityNav achieves both scalability and responsiveness for large-scale multi-vehicle navigation.

To align local decisions with global optimization goals, CityNav introduces a cooperative reasoning optimization mechanism that jointly trains global and local LLM agents to balance per-vehicle efficiency and network-wide mobility.
During training, agents interact with the traffic environment and receive real-time feedback on routing efficiency and congestion. Each agent is first optimized with an individual reward that reflects its task-specific objective (\ie minimizing network congestion for the global traffic allocation and maximizing regional mobility for the local navigation). Then, both agents are jointly optimized with a shared network-wide reward, promoting coordination toward overall traffic efficiency.
This hierarchical dual-reward optimization mechanism enables CityNav to achieve globally efficient yet locally adaptive navigation across large-scale urban road networks.

The key contributions of this work are threefold: 
(1) We propose CityNav, the first LLM-based multi-agent framework designed for city-scale dynamic navigation. By leveraging hierarchical coordination, CityNav enables scalable and adaptive routing across large road networks.
(2) We design a cooperative reasoning optimization mechanism that explicitly aligns local decisions with global objectives. This approach fosters coordination among hierarchical agents, balancing individual performance with network-wide efficiency.
(3) We conduct extensive experiments on four real-world road networks of varying scales (up to 1.6 million roads and 430,000 intersections) and traffic datasets, benchmarking CityNav against nine classical path search and RL-based methods. Results show consistent improvements in effectiveness, scalability, and computational efficiency.

%% file: sections/preliminary.tex
\section{Preliminary}

Let the urban road network be represented as a directed graph $G = (V, E)$, where $V$ is the set of intersections and $E$ is the set of directed road segments. Traffic dynamics evolve over this network as vehicles move, interact, and affect congestion levels.

\begin{definition}[Multi-Vehicle Dynamic Navigation]
    The multi-vehicle dynamic navigation is formulated as a partially observable Markov decision process (POMDP), defined by a tuple $\langle S, A, O, R, N \rangle$:
\begin{itemize}[leftmargin=0.5cm]
    \item \textbf{Traffic state}: $s$ is the traffic state, describing the joint condition of the road network;
    \item \textbf{Observation space}: $o \subseteq s$ represents the local observation of navigation control agents;
    \item \textbf{Routing plan}: $P$ is the routing plan of a vehicle, corresponding to the trajectory of traversing regions or roads;
    \item \textbf{Reward function}: $r=f_{\text{reward}}(s, \mathbf{P})$ is the reward function for navigation control agents, where $\mathbf{P} = \{P_0, \dots, P_N\}$ is the collection of routing plans of controlled vehicles.
\end{itemize}
Each vehicle travels from an origin node $v_1 \in V$ to a destination node $v_d \in V$ through a sequence of road segments while minimizing travel delay and avoiding congestion.
\end{definition}

%% file: sections/method.tex
\section{LLM for Multi-Vehicle Dynamic Navigation}

\begin{figure}
    \centering
    \includegraphics[width=\linewidth]{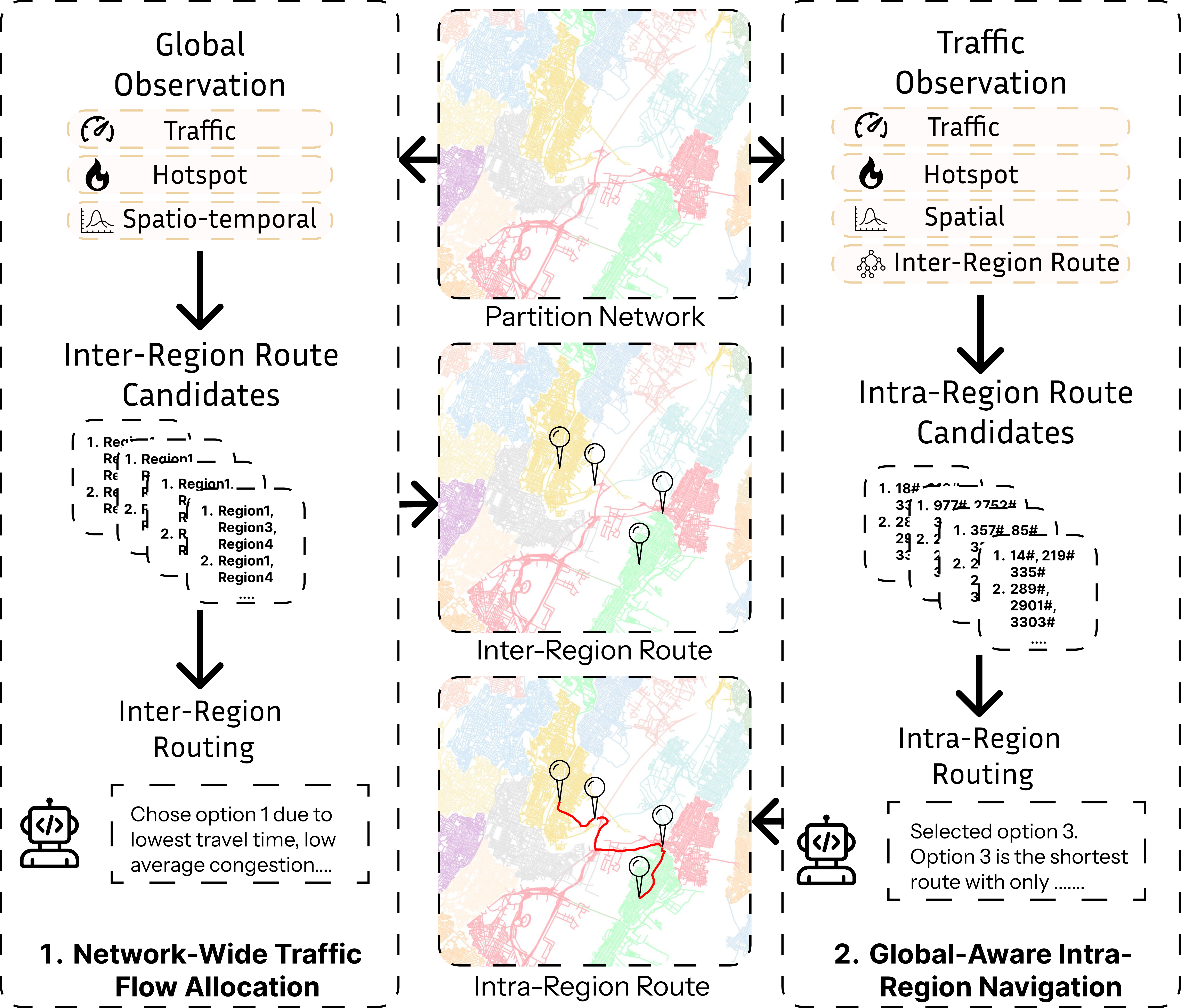}
    \caption{Workflow of CityNav framework.}
    \label{fig:archi}
\vspace{-10pt}
\end{figure}

As illustrated in Figure~\ref{fig:archi}, the proposed \textit{CityNav} framework introduces a hierarchical reasoning architecture that integrates global coordination with local adaptability to address large-scale multi-vehicle navigation. (1) At the upper level, a global traffic flow allocation agent oversees network-wide traffic conditions and dynamically formulates high-level routing strategies to balance flow distribution across different regions. (2) At the lower level, a set of local navigation agents operates within their assigned regions, where each agent generates fine-grained routing decisions tailored to local traffic dynamics. Guided by the global agent’s strategic directives, local agents perform contextual reasoning to adapt to real-time variations, collaboratively ensuring efficient and congestion-aware vehicle navigation throughout the entire network.

\subsection{Network-Wide Traffic Flow Allocation}

At the upper level of CityNav, the global traffic allocation agent serves as the strategic coordinator that regulates network-wide traffic distribution. It aims to construct macro-level routing plans to avoid potential congestion. Before the operational loop begins, the entire road network is partitioned into a set of non-overlapping regions $\mathbf{z} = \{z_1, z_2, \dots, z_K\}$ using the Louvain community detection algorithm \cite{blondel2008fast}, which groups spatially and topologically cohesive areas while preserving strong intra-region connectivity.

\subsubsection{Observation Construction}
At each decision step, the traffic allocation agent receives an observation $o = \{o_{\text{traffic}}, o_{\text{hot}}, o_{\text{st}}\}$ composed of three categories of real-time features that summarize global traffic conditions: 
(1) \textit{Traffic conditions} ($o_{\text{traffic}}$): For each region $z_k \in \mathbf{z}$, we define a traffic descriptor:
\begin{align}
    o^{z_k}_{\text{traffic}} = [\text{cong}(z_k), \text{occ}(z_k), \bar{\tau}(z_k)],
\end{align}
where $\text{cong}(z_k)$, $\text{occ}(z_k)$, and $\bar{\tau}(z_k)$ denote the congestion level, occupancy rate, and average travel time within region $z_k$. 
(2) \textit{Hotspot indicator} ($o_{\text{hot}}$): The current demand intensity of $z_k$ is given by:
\begin{align}
   o_{hot}^{z_k} = \frac{N^k_{\text{active}}}{N_{\text{total}}},
\end{align}
where $N^k_{\text{active}}$ is the number of active vehicles whose routing plan include $z_k$.
(3) \textit{Spatio-temporal contexts} ($o_{\text{st}}$): Environmental context is captured by the set of congested edges $E_{\text{cong}}$ and the current time index $t$, allowing the LLM to reason with temporal traffic priors and known spatial traffic bottlenecks.

\subsubsection{Network-Wide Traffic Distribution}
Given the observation $o$, the vehicle’s origin region $z_1$, and destination region $z_d$, we construct a traffic-aware prompt comprising: the current traffic observation $o$, the origin–destination pair $\langle z_1, z_d \rangle$, and a set of pre-screened, physically traversable candidate inter-region routes $\mathbf{P}_{\text{global}} = \{P_1, \dots, P_M\}$. Each route $P_m = \{z_1, \dots, z_d\} \subseteq \mathbf{z}$ represents an ordered sequence of regions connecting $z_1$ to $z_d$. Formally, the prompt is written as:
\begin{align}
X_{\text{global}}  = \text{Prompt}(o, z_1, z_d, \mathbf{P}_{\text{global}}).
\end{align}
The allocator’s decision-making then proceeds in two stages, inspired by the ReAct framework~\cite{yao2023react}. First, during the reasoning phase, it analyzes the observation $o$, predicts emerging congestion patterns, and leverages its pre-trained knowledge of typical spatio-temporal dependencies to evaluate each candidate plan.
Based on the reasoning, the allocation agent then selects the optimal route that minimizes expected travel delay and avoids congestion:
\begin{align}
P_{\text{global}}^* = \pi_{\text{global}}^{\text{routing}}\big(\mathbf{P}; \pi_{\text{global}}^{\text{reason}}(X_{\text{global}})\big),
\end{align}
This two-stage process enables the global allocator to generate strategic, traffic-aware routing plans that are both interpretable and grounded in a comprehensive understanding of current and anticipated network-wide traffic conditions.

Operating at the lower level of CityNav, decentralized local navigation agents serve as tactical executors that translate the high-level strategic directives from the global traffic allocation agent into concrete, real-time routing plans within their assigned regions. The navigation is activated whenever a vehicle enters the assigned region of the agent, requesting an immediate control signal on which road segment to traverse, while remaining aligned with the global inter-region routing plan $P_{\text{global}}^*$.

\subsubsection{Observation Construction}

To make context- and traffic-aware decisions, we construct a road segment-level observation $o' = \{o'_{\text{traffic}}, o'_{\text{hot}}, o'_{\text{spatial}}\}$ by aggregating three categories of local features: 
(1) \textit{Traffic conditions} ($o'_{\text{traffic}}$): For each road segment $e$ within the region, the traffic condition is described by:
\begin{align}
    o'^{e}_{\text{traffic}} = [\text{cong}(e), \text{occ}(e), 
    \hat{\tau}(e)],
\end{align}
where $\text{cong}(e)$ is the congestion level, $\text{occ}(e)$ is the occupancy rate of the segment, and $\hat{\tau}(e)$ is the flow-free travel time (\ie travel time without external delays such as congestion or red-light waiting).
(2) \textit{Hotspot indicator} ($o'_{\text{hot}}$): The immediate demand for boundary edges $E_{\text{bound}}$ connecting the next target region $z_{k+1}$ is quantified by $N^\text{bound}_{\text{active}}$, the number of vehicles currently routed toward road $E_{\text{bound}}$.
(3) \textit{Spatial contexts} ($o'_{\text{spatio}}$): This encodes the vehicle’s current location $e_i$ and the next region $z_{k+1}$ in the global inter-region routing plan $P^*_{\text{global}}$, providing critical guidance to ensure local routing decisions are aligned with the network-wide strategy.

\subsubsection{Global-Aware Route Planning}

Firstly, we perform candidate path pre-screening, where all feasible paths from the vehicle’s current location to the boundary roads are ranked by estimated free-flow travel time. The top ten paths form the candidate micro-routing plan set $\mathbf{P}_{\text{local}}$, focusing the LLM agent’s decision-making on the most promising options. Given the local observation $o'$, the target region $z_{k+1}$, and the candidate routing plan set $\mathbf{P}_{\text{local}}$, the navigation agent executes a two-stage ReAct process, informed by the global intra-region plan $P_{\text{global}}^*$:
\begin{align}
    & X_{\text{local}} = \text{Prompt}(o', z_{k+1}, \mathbf{P}_{\text{local}}), \\
    & P^*_{\text{local}} = \pi^{\text{routing}}_{\text{local}}(\mathbf{P}_{\text{local}};\langle \pi^{\text{reason}}_{\text{local}}(X_{\text{local}}), P_{\text{global}}^*\rangle).
\end{align}
This hierarchical, context- and globally-aware routing mechanism ensures that intra-region decisions are both locally responsive and strategically aligned, enabling interpretable and efficient navigation across the entire traffic network.

\section{Cooperative Reasoning Optimization}

While the hierarchical CityNav framework effectively reduces the complexity of network-wide vehicle navigation, off-the-shelf LLMs are not explicitly designed for coordinated, top-down vehicle navigation control. To address this, we introduce a cooperative reasoning optimization mechanism that enables agents to learn from the collective impact of their mutual routing decisions. Within this paradigm, the global traffic allocator and decentralized local navigation agents are jointly trained to reason toward a shared system-level objective, ensuring that local routing decisions remain globally consistent. This design allows CityNav to maintain network-wide efficiency while preserving the flexibility under real-time dynamics.

\subsection{Coordination Reward Construction}

To facilitate cooperation across hierarchical agents, we design a dual-reward mechanism composed of an individual reward that promotes per-vehicle efficiency and a shared reward that aligns all agents toward global traffic optimization.

The individual reward quantifies per-vehicle efficiency while maintaining fairness across routes with different traffic contexts:
\begin{align}
    r_{\text{ind}} = \frac{\tau}{\tau + \hat{\tau}} - \lambda \cdot N_{\text{idle}},
\end{align}
where $\tau$ is the actual travel time, $\hat{\tau}$ is the flow-free travel time, $N_{\text{idle}}$ is the number of idle steps, and $\lambda$ controls the penalty for prolonged idling. This normalization mitigates bias toward short routes and encourages smooth movement by penalizing unnecessary stops under dynamic congestion.

To enable cooperative reasoning between two hierarchical agents, we introduce a shared reward that reflects network-wide mobility:
\begin{align}
    r_{\text{share}} = - \frac{1}{|P^*_{\text{global}}|} \sum_{z \in P^*_{\text{global}}} \bar{\tau}(z),
\end{align}
where $P^*_{\text{global}} = \{z_1, \dots, z_d\}$ denotes the sequence of regions in the global routing plan, and $\bar{\tau}(z)$ is the average travel time of vehicles traversing region $z$. This formulation provides a unified optimization signal: the global traffic allocation agent learns to minimize aggregated regional delays, while local navigation agents adapt their routing to support the global plan.

\subsection{Bi-Level Reasoning Optimization}

Building upon this reward structure, we develop a bi-level reasoning optimization mechanism that jointly post-trains the global traffic allocation and local navigation agents toward a cooperative equilibrium. This framework ensures that reasoning processes across both levels are optimized to balance per-vehicle efficiency and network-wide coordination.

At each training iteration, for each agent, we sample $N$ reasoning processes under the traffic state $s$:
\begin{align}
\mathcal{D} = \{y_1, y_2, \ldots, y_{N}\} \sim \pi_{\theta}(\cdot \mid s),
\end{align}
where each reasoning process $y_i$ represents a full reasoning–action sequence generated by policy $\pi_{\theta}$.
Each reasoning process receives a scalar reward combining individual and overall objectives:
\begin{align}
r_i = \alpha \, r_{\text{ind}}^i + (1 - \alpha) \, r_{\text{share}}^i,
\end{align}
where $\alpha \in [0,1]$ controls the trade-off between per-vehicle efficiency and network-wide coordination.

Inspired by \cite{wan2025rema}, we adopt a multi-agent group reinforcement policy optimization (GRPO) \cite{guo2025deepseek} process to jointly optimize the two hierarchical agents. To stabilize optimization, we compute a group-normalized advantage without the critic network:
\begin{align}
A_i = r_i - \bar{r}, \quad \text{where} \quad 
\bar{r} = \frac{1}{N} \sum_{j=1}^{N} r_j.
\end{align}
The positive advantage indicates the reasoning process outperforms the group average, directly guiding policy improvement. The moving average of $\bar{r}$ is optionally used to smooth variance across training steps. For each reasoning process, the clipped surrogate loss is then defined as:
\begin{align}
L_{\text{clip}}^i(\theta) = \min \Big( \frac{\pi_{\theta}(y_i \mid s)}{\pi_{\theta_{\text{ref}}}(y_i \mid s)}A_i, \text{clip}(\frac{\pi_{\theta}(y_i \mid s)}{\pi_{\theta_{\text{ref}}}(y_i \mid s)},1-\epsilon,1+\epsilon), A_i \Big),
\end{align}
where $\epsilon$ is the clipping parameter that stabilizes policy updates. The overall cooperative GRPO objective jointly optimizes both agents’ parameters $\Theta=\{\theta_\text{global}, \theta_{\text{local}}\}$ while enforcing alignment through a KL penalty:
\begin{align}
\max_{\Theta} \; 
\mathbb{E}_{i \sim \mathcal{D}} \!\left[
\sum_{\ell \in \{\text{global}, \text{local}\}} \!\left(
\frac{1}{N} \sum_{i=1}^{N} L_{\text{clip}}^i(\theta_\ell)
- \beta \, \mathbb{D}_{\text{KL}}\!\left(\pi_{\theta_\ell} \,\|\, \pi_{\theta_{\text{ref}}^\ell}\right)
\right)
\right],
\end{align}
where $\Theta = \{\theta_\text{global}, \theta_\text{local}\}$ are model parameters of global and local agents, and $\beta$ is the KL coefficient controlling update stability.

During training, both agents are optimized asynchronously using the AdamW optimizer with progressive learning rate schedules. Loss masking is applied to exclude observation tokens from gradient computation, along with mixed-precision (AMP) training and gradient clipping. Through this cooperative bi-level optimization, CityNav progressively aligns the reasoning and policies of hierarchical LLM agents, achieving locally adaptive yet globally consistent network-wide vehicle navigation.

%% file: sections/exp.tex





\section{Experiments}
In this section, we conduct extensive experiments to evaluate our proposed CityNav by answering the following research questions:
\begin{itemize}[leftmargin=0.5cm]
    \item How is the \textbf{effectiveness} of CityNav compared with traditional path search algorithms and RL-based methods?
    \item How is the \textbf{scalability} ability of CityNav across different scales of road networks and traffic volumes?
    \item How is the \textbf{generalization} of CityNav in unseen road networks and traffic contexts?
\end{itemize}

\subsection{Experimental Settings}

\begin{table}[t] 
\setlength{\tabcolsep}{2pt}
  \centering 
  \caption{Statistics of datasets.} 
  \label{tab:my_data_stats} 
  \small
  \resizebox{\columnwidth}{!}{
  \begin{tabular}{l l c r r r r}
    \toprule
    \multirow{2}{*}{\textbf{Dataset}} & \multirow{2}{*}{\textbf{Intersections/Roads}} & \multirow{2}{*}{\textbf{Vehicles}} & \multicolumn{4}{c}{\textbf{Arrival rate (vehicles/5min)}} \\
    \cmidrule(lr){4-7} 
                     &                              &                   & \textbf{Mean}   & \textbf{Std}    & \textbf{Max}    & \textbf{Min}    \\
    \midrule
    NYC            & 429,581/1,639,481            & 1,038,608         & 3,570.28        & 1,359.76        & 10,675          & 999             \\
    Manhattan                & 18,240/74,979                & 24,880            & 86.39           & 80.04           & 332             & 12              \\
    UES & 3,955/11,134                 & 2,000             & 27.03           & 2.16            & 28              & 9               \\
    Chicago                  & 249,511/978,799              & 79,342            & 596.56          & 9.42            & 600             & 490             \\
    \bottomrule
  \end{tabular}}
\vspace{-10pt}
\end{table}

\subsubsection{Datasets}
We conduct our experimental evaluation on four real-world urban traffic datasets that differ in both spatial scale and network topology. The origin–destination (OD) data for controlled vehicles are derived from the New York City and Chicago Taxi datasets \cite{nyc_tlc_green_taxi_2015,chicago_taxi_trips_2013_2023}, which capture fine-grained mobility demand patterns across time and space. A randomly selected 2\% of trips is designated as optimization targets managed by our framework. To faithfully reconstruct realistic city-scale traffic conditions, the remaining 98\% of trips, along with additional background vehicle volumes, are calibrated using official mobility surveys \cite{nyc_citywide_mobility_survey}. Following \cite{zhao2024origin}, A gravity-based model \cite{NBERw16576} is deployed for inter-region OD demand generation, where the background traffic flow between two regions is proportional to their activity intensities.

\begin{itemize}[leftmargin=0.5cm]
    \item \textbf{New York City (NYC)}: The complete road network of New York City serves as the primary experimental environment. It comprises 429,581 intersections and captures diverse traffic dynamics across urban, suburban, and peripheral areas.
    \item \textbf{Manhattan}: The Manhattan sub-network, consisting of 18,240 intersections, represents a dense urban core characterized by recurrent congestion, grid-like topology, and complex interactions between arterial and local roads.
    \item \textbf{Upper East Side (UES)}:  
    The Upper East Side, a compact subregion within Manhattan, NYC, containing 3,955 intersections, provides a small-scale setting for evaluating localized navigation and coordination behaviors.
    \item \textbf{Chicago}: Chicago, with 249,511 intersections, is used to evaluate the zero-shot generalization ability of our framework across distinct city layouts and mobility distributions.
\end{itemize}

\subsubsection{Data Anonymization}

The taxi trip data used in our experiments is sourced from publicly available NYC and Chicago Open Data \cite{nyc_opendata_portal,chicago_data_portal}. The dataset is anonymized and does not contain any personally identifiable information regarding drivers or passengers.

\subsubsection{Environment Settings}

All experiments are conducted using SUMO \cite{SUMO2018}, a widely adopted open-source microscopic traffic simulator that enables high-fidelity modeling of vehicular dynamics and network-level interactions. SUMO supports large-scale simulations with realistic mobility and provides an interactive visualization interface for monitoring traffic states. It offers diverse APIs for retrieving traffic features and implementing external control policies.

\subsection{Metrics}

Following prior studies \cite{lai2024llmlightlargelanguagemodels,yin2025cooperative,Luo_Wang_Zhang_Yuan_Li_2023}, we evaluate the performance of baseline methods using four key traffic efficiency metrics: average travel time (ATT), average waiting time (AWT), average delay time (ADT), and throughput (TP). These metrics jointly capture both individual-level mobility efficiency and overall network throughput.

\begin{itemize}[leftmargin=0.5cm]
    \item \textbf{Throughput (TP)}:
    The total number of vehicles that successfully reach their destinations within the simulation period. Higher values indicate better overall throughput.
    \item \textbf{Average Travel Time (ATT)}:
    The mean duration of all completed trips, measured from origin to destination. Lower values indicate more efficient routing.
    \item \textbf{Average Waiting Time (AWT)}:
    The mean total time vehicles remain stationary in the network (\eg at intersections or due to congestion). Lower values are better.
    \item \textbf{Average Delay Time (ADT)}:
    The average deviation between a vehicle’s actual travel time and its theoretical free-flow travel time. Lower values reflect smoother network operations.
\end{itemize}

\subsection{Compared Approaches}

To comprehensively evaluate the effectiveness of CityNav, we benchmark it against nine representative baselines. For path search algorithms, we select: \textbf{Dijkstra} \cite{dijkstra2022note} serves as the classical deterministic baseline, computing globally optimal shortest paths based on static network topology and road length; \textbf{MinDits} adopts a greedy strategy that selects the next-hop road closest to the destination in terms of network distance. \textbf{MinLat} follows a similar greedy heuristic but chooses the next-hop road with the smallest estimated travel latency based on predicted traffic conditions using ARIMA \cite{box1976analysis}; \textbf{SBP}~\cite{ijcai2021p503}, which optimizes continuous route combinations; \textbf{DyAM}~\cite{LECLERCQ2021118}, which enforces routing through dynamic avoidance maps; and \textbf{S-DTA}~\cite{https://doi.org/10.1111/mice.12577}, a meta-heuristic framework for simulation-based dynamic traffic assignment. For RL-based methods, we include: \textbf{XRouting}~\cite{wang2022xrouting}, an early deep reinforcement routing method; \textbf{Adaptation-Navigation}~\cite{10.1145/3557915.3561005}, which emphasizes adaptive policy learning under non-stationary traffic conditions; and \textbf{AlphaRoute}~\cite{Luo_Wang_Zhang_Yuan_Li_2023}, the SOTA multi-agent coordination framework.

\subsection{Model Settings}

We adopt Qwen-3-8B \cite{yang2025qwen3technicalreport} as the backbone LLM, which offers strong performance across diverse tasks while maintaining a lightweight model size. To ensure stable and reproducible training, a consistent set of hyperparameters is used across all experiments. A conservative learning rate of $1 \times 10^{-4}$ is applied, with a replay buffer capacity of 10,000 experiences. Training follows a mini-batch strategy with a base batch size of 4 and 8 steps of gradient accumulation. For the LLM configuration, we set \texttt{top\_p} as 1.0 and the \texttt{temperature} to 0.1. Low-Rank Adaptation (LoRA) fine-tuning is employed with a rank of 8 and a scaling factor $\alpha = 16$. Post-training data is collected from interactive simulations on the NYC dataset.

\begin{table*}[t]
  \centering
  \caption{Overall performances of CityNav and traditional methods on New York City, Manhattan, and Upper East Side datasets. The symbol “–” indicates that the corresponding model failed to converge during training on the given dataset.}
  \resizebox{\linewidth}{!}{
  \begin{tabular}{l|cccc|cccc|cccc}
    \toprule
    \multirow{2}{*}{\textbf{Methods}} & \multicolumn{4}{c|}{\textbf{New York City}} & \multicolumn{4}{c|}{\textbf{Manhattan}} & \multicolumn{4}{c}{\textbf{Upper East Side}} \\
    \cmidrule{2-13}
    & \textbf{TP $\uparrow$} & \textbf{ATT $\downarrow$} & \textbf{AWT $\downarrow$} & \textbf{ADT $\downarrow$} & \textbf{TP $\uparrow$} & \textbf{ATT $\downarrow$} & \textbf{AWT $\downarrow$} & \textbf{ADT $\downarrow$} & \textbf{TP $\uparrow$} & \textbf{ATT $\downarrow$} & \textbf{AWT $\downarrow$} & \textbf{ADT $\downarrow$} \\
    \midrule
    \multicolumn{13}{c}{\textbf{Path Search Algorithms}} \\
    \midrule
    Dijkstra & 645 & 703.83 & 98.88 & 36785.70 & 378 & 1315.14 & 23.93 & 1002.03 & 40 & 274.82 & 42.10 & 167.04 \\
    MinDits & 632 & 810.48 & 98.89 & 36889.75 & 356 & 1370.99 & 24.88 & 1021.35 & 40 & 276.68 & 43.23 & 186.92 \\
    MinLat & 632 & 810.47 & 98.89 & 36889.79 & 356 & 1370.79 & 24.58 & 1096.39 & 40 & 276.68 & 43.23 & 176.20 \\
    SBP & 488 & 10874.38 & 91.23 & 20480.81 & 193 & 1140.50 & 24.27 & 789.99 & 40 & 255.35 & 39.80 & 137.70 \\
    DyAM & 778 & 10759.85 & 92.57 & 20332.25 & 262 & 1072.23 & \textbf{23.29} & 693.49 & 40 & 275.87 & 37.08 & 145.74 \\
    S-DTA & 758 & \textbf{645.43} & 99.19 & 35669.84 & 257 & 1033.98 & 49.44 & 730.36 & 40 & 255.08 & 107.23 & 156.49 \\
    \midrule
    \multicolumn{13}{c}{\textbf{RL-based Methods}} \\
    \midrule
    Xrouting & - & - & - & - & 217 & 1360.48 & 33.37 & 1091.13 & 40 & 203.24 & 43.23 & 107.65 \\
    Adaptation-Navigation & - & - & - & - & 241 & 1599.33 & 36.05 & 1255.24 & 40 & 230.89 & 21.47 & 134.39 \\
    Alpharoute & - & - & - & - & 225 & 1381.64 & 48.85 & 1083.87 & 40 & \textbf{199.78} & \textbf{7.78} & \textbf{104.92} \\
    \midrule
    CityNav & \textbf{2952} & 2771.21 & \textbf{89.34} & \textbf{4207.46} & \textbf{393} & \textbf{667.99} & 30.09 & \textbf{357.79} & \textbf{40} & 257.52 & 38.93 & 135.29 \\
    \bottomrule
  \end{tabular}
  }
  \label{tab:BaselineComparison}
\end{table*}

\subsection{Comparison Between the Traditional Methods and CityNav (RQ1 \& RQ2)}

We first conduct experiments on three different scales of real-world traffic scenarios, including NYC, Manhattan, and Upper East Side, to evaluate the effectiveness and scalability of our proposed framework. The comparison between traditional and learning-based approaches is summarized in Table~\ref{tab:BaselineComparison}.

CityNav demonstrates significant advantages in diverse scales of road networks, achieving the highest throughput with consistently low ATT, AWT, and ADT. On the New York City dataset, the largest and most complex scenario, CityNav successfully completes nearly three thousand trips, achieving more than four times the throughput of the best competing approach while maintaining efficient travel and delay times. In contrast, the path-search algorithms (\eg Dijkstra, MinDits, and S-DTA) suffer from severe congestion, completing only about 21\% of the throughput achieved by CityNav. Although some of these methods appear to report lower ATT, most vehicles remain stuck in traffic congestion and fail to reach their destinations, resulting in relatively low ATT that reflects only a small subset of successful trips. Notably, no RL-based method converged in this city-scale navigation scenario, highlighting the challenges of trial-and-error training such models under large, dynamically coupled traffic environments.

On the smaller Upper East Side network, recent path search algorithms (\eg SBP and S-DTA) and RL-based methods (\eg Alpharoute) achieve strong performance, benefiting from manageable state and action spaces. However, as the network scale increases to the full Manhattan and New York City settings, their performance deteriorates sharply or becomes infeasible to train due to the exponential growth of environment complexity. Conversely, CityNav maintains robust scalability, effectively managing network-wide coordination and congestion even under diverse traffic settings.

Overall, these results confirm that CityNav delivers globally efficient and congestion-resilient traffic management across varying network scales. Our proposed method outperforms both path search and RL-based navigation baselines while maintaining strong scalability to complex, real-world traffic environments.

\subsection{Demand Scalability Comparison (RQ2)}

A key requirement for any city-scale navigation framework is the ability to scale effectively under increasing traffic demand. To examine this property, we evaluate all methods on the New York City dataset by varying the proportion of optimized trips from 2\% to 10\% of total origin–destination (OD) demand during the morning peak hours (8 AM–12 PM). This setting simulates a progressively denser traffic environment, where the number of vehicles subject to active coordination rises fivefold. As shown in Table~\ref{tab:ratio_performance}, higher demand naturally leads to increased congestion, reflected by longer average travel time (ATT), average waiting time (AWT), and average delay time (ADT) across all systems.

From the results, CityNav demonstrates remarkable robustness and scalability. When the proportion of optimized vehicles increases fivefold, throughput (TP) grows steadily from 1,022 to 4,294 while maintaining moderate increases in ATT, AWT, and ADT relative to the traffic surge. This stability suggests that CityNav’s cooperative, bi-level reasoning effectively coordinates vehicles to mitigate localized congestion and sustain network-wide flow efficiency. The results confirm that CityNav generalizes well across different traffic intensities, preserving both global coordination stability and local adaptability even in heavy-demand scenarios.

\begin{table}[t]
\setlength{\tabcolsep}{8pt}
  \centering
  \caption{Performance in different scales of traffic demand.}
  \resizebox{\linewidth}{!}{
  \begin{tabular}{l|l|cccc}
    \toprule
    \multirow{2}{*}{\textbf{Datasets}} & \multirow{2}{*}{\textbf{Ratio}} & \multicolumn{4}{c}{\textbf{Metrics}} \\
    \cmidrule(l){3-6} 
    & & \textbf{TP} & \textbf{ATT} & \textbf{AWT} & \textbf{ADT} \\
    \midrule
    \multirow{3}{*}{New York} & 2\% & 1022 & 1412.64 & 35.38 & 2019.56  \\
    & 5\% &  2489& 1975.57& 47.77&2722.46\\
    & 10\% & 4294& 2922.81& 68.43&3871.39 \\
    \bottomrule
  \end{tabular}
  }
  \label{tab:ratio_performance}
\vspace{-10pt}
\end{table}

\subsection{Generalization Comparison (RQ3)}

To evaluate the model’s ability to generalize beyond its training environment, we conducted a zero-shot transfer experiment. Models trained on the New York City road network were directly deployed on the Chicago network, which differs substantially in both topology and traffic dynamics. This setup tests whether each method learns transferable routing principles or merely overfits to the specific spatial and flow characteristics of the training environment. The results are summarized in Table~\ref{tab:zero_shot}.

CityNav demonstrates strong generalization capability, achieving a throughput of 1,073 successfully routed vehicles, approximately twice that of the best-performing path search algorithm, while maintaining considerably lower ATT, AWT, and ADT. This shows that CityNav effectively transfers its learned coordination strategies to an unseen city without retraining, preserving both routing efficiency and network stability under new conditions. In contrast, the path-search algorithms (\eg Dijkstra, S-DTA, and MinDits) suffer from severe congestion, resulting in a drastic reduction in throughput as most vehicles become trapped in traffic and only a few short trips complete successfully. Overall, these results confirm that CityNav not only scales efficiently with increasing traffic volumes but also generalizes robustly across heterogeneous urban topologies, achieving consistent, congestion-resilient performance even in unseen road network encironments.

\begin{table}[t]
  \centering
  \caption{Zero-shot experiment on the Chicago dataset.}
  \resizebox{\linewidth}{!}{
  \begin{tabular}{l|l|cccc}
    \toprule
    \multirow{2}{*}{\textbf{Datasets}} & \multirow{2}{*}{\textbf{Methods}} & \multicolumn{4}{c}{\textbf{Metrics}} \\
    \cmidrule(l){3-6} 
    & & \textbf{TP} & \textbf{ATT} & \textbf{AWT} & \textbf{ADT}  \\
    \midrule
    \multirow{7}{*}{\textbf{Chicago}} & Dijkstra & 595 & \textbf{419.57} & 78.48 & 14388.15  \\
    & MinDits & 683 & 460.92 & 98.95 & 14265.66  \\
    & MinLat & 683 & 482.25 & 98.96 & 14288.87  \\
    & SBP & 758 & 4897.45 & 60.33 & 12859.95 \\
    & DyAM & 788 & 4996.85 & 76.52 & 12796.56  \\
    & S-DTA & 655 & 475.31 & 68.42 & 16745.71  \\
    & CityNav & \textbf{1073} & 1609.40 & \textbf{58.56} & \textbf{2561.78}  \\
    \bottomrule
  \end{tabular}
  }
  \label{tab:zero_shot}
\vspace{-5pt}
\end{table}

\subsection{Comparison with Leading LLMs}

\begin{figure}
    \centering
    \includegraphics[width=\linewidth]{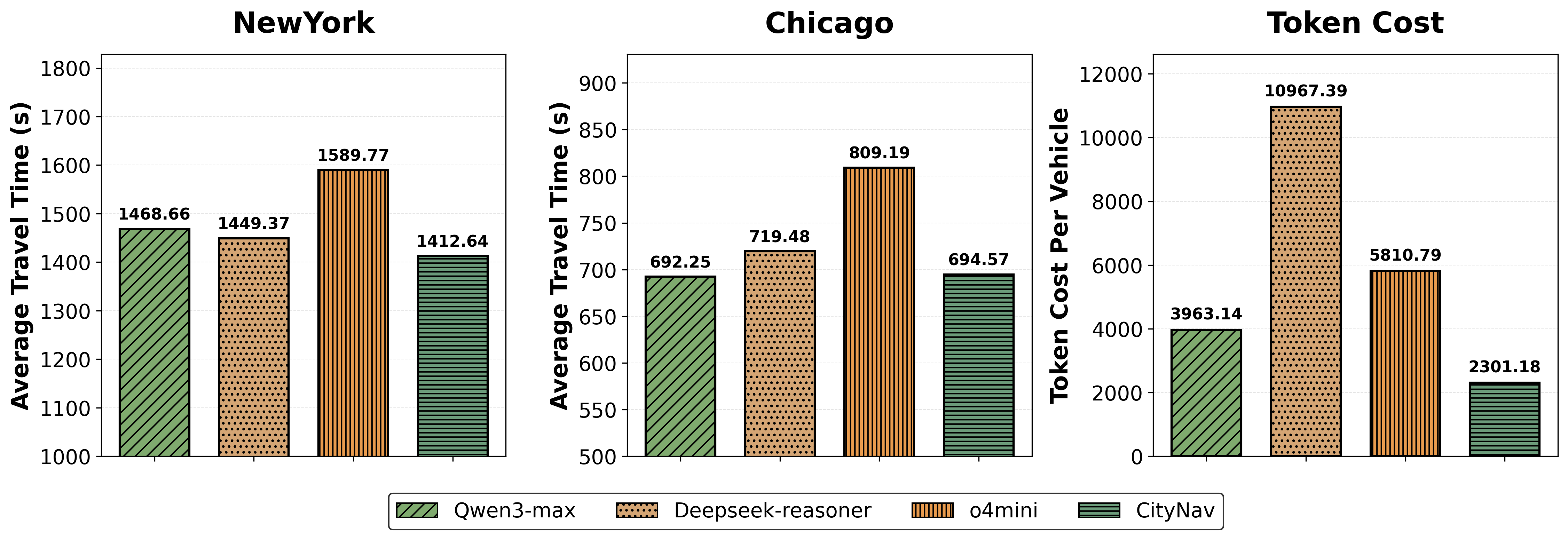}
    \caption{Comparison between leading LLMs and CityNav.}
    \label{fig:LLMs}
    \vspace{-5pt}
\end{figure}

This final experiment examines the significance of CityNav’s domain-specific fine-tuning by replacing its backbone model with several state-of-the-art general-purpose LLMs, including Qwen3-Max (over 1T parameters), DeepSeek-R1 (761B), and o4-mini (close-source). Evaluations were conducted on both the NYC dataset for in-domain evaluation and the Chicago dataset for zero-shot generalization examination during the morning rush hour (8 AM–12 PM). The results are presented in Figure~\ref{fig:LLMs}.

Despite being built upon a lightweight 8B-parameter backbone, CityNav consistently achieves the lowest ATT across both datasets, matching or surpassing much larger off-the-shelf LLMs. This superior performance arises from CityNav’s bi-level cooperative fine-tuning, which enable coordinated and adaptive reasoning under dynamic congestion conditions. Moreover, CityNav attains these results with the least token usage (58.06\% of the tokens used by Qwen3-Max), owing to its hierarchical reasoning structure that minimizes redundant inference and communication among agents. This exceptional efficiency–performance balance underscores CityNav’s scalability and practicality. This demonstrates that our proposed targeted domain optimization effectively surpass generic large-scale reasoning models in both effectiveness and computational efficiency in real-world depolyment scenarios.

\subsection{Ablation Study}

To evaluate the contributions of CityNav’s key components, we conducted ablation experiments on two primary variants: (1) CityNav w/o hierarchy, where the multi-level agent structure is collapsed into a single centralized agent responsible for all routing decisions, testing the necessity of hierarchical decomposition for handling city-scale state spaces; and (2) CityNav w/o RL, which retains the hierarchy but omits the cooperative reasoning optimization, isolating the effect of reinforcement learning on coordination. All three versions, including the full CityNav, were evaluated on the NYC dataset from 0 AM to 12 PM. The results are summarized in Table~\ref{tab:Abalation}.

The results demonstrate that both the hierarchical framework and RL-based reasoning optimization are essential for CityNav’s performance. Without the hierarchy, the centralized agent fails to efficiently capture network-wide traffic conditions while managing individual vehicle routing, leading to severe bottlenecks and drastically reduced throughput. Removing RL weakens the coordination between global traffic allocation and local navigation agents, resulting in suboptimal routing and higher travel times. In contrast, the full CityNav model effectively balances per-vehicle efficiency with global traffic mobility, achieving substantially lower ATT, AWT, and ADT while completing nearly twice as many trips. These findings confirm that CityNav can robustly handle city-scale traffic optimization with efficient multi-agent coordination.

\begin{table}[t]
  \centering
  \caption{Ablation tests.}
  \resizebox{\linewidth}{!}{
  \begin{tabular}{l|l|cccc}
    \toprule
    \multirow{2}{*}{\textbf{Datasets}} & \multirow{2}{*}{\textbf{Methods}} & \multicolumn{4}{c}{\textbf{Metrics}} \\
    \cmidrule(l){3-6} 
    & & \textbf{TP} & \textbf{ATT} & \textbf{AWT} & \textbf{ADT}  \\
    \midrule
    \multirow{3}{*}{NewYork} & w/o RL & 1883 & 6762.54 & \textbf{53.25} & 9324.45  \\
    & w/o hierarchy & 493 & 9983.78 & 93.37 & 18853.75  \\
    & CityNav & \textbf{2952} & \textbf{2771.21} & 89.34 & \textbf{4207.46}  \\
    \bottomrule
  \end{tabular}
  }
  \label{tab:Abalation}
\vspace{-5pt}
\end{table}

\begin{figure}[t]
    \centering
    \includegraphics[width=\linewidth]{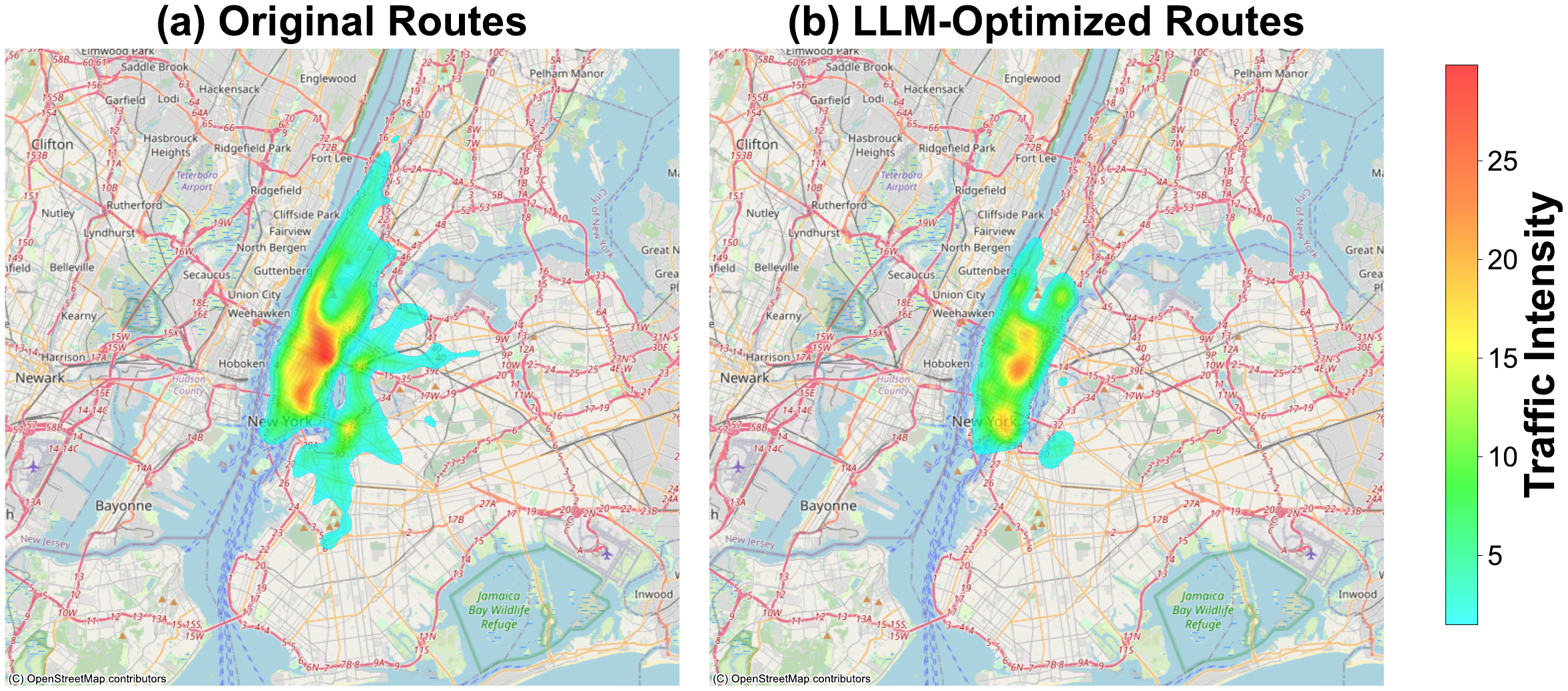}
    \caption{Traffic coordinated by Dijkstra and CityNav.}
    \label{fig:Casestudy}
\end{figure}

\subsection{Case Study}

To provide a macroscopic and intuitive view of CityNav’s impact on urban traffic, we conducted a case study visualizing traffic distribution across the New York City road network. In Figure~\ref{fig:Casestudy}, we show traffic heatmaps comparing routing coordinated by Dijkstra versus optimized by CityNav, with color intensity indicating the number of vehicle passes on each road segment.

The comparison demonstrates that CityNav significantly mitigates congestion across the city. Before deployment, heavy traffic accumulations were concentrated throughout central areas. After deploying CityNav, congestion in the hotspots (\ie regions in red) is markedly reduced, while traffic flow in the rest of the network becomes smoother and more evenly distributed. Overall, CityNav promotes a more balanced and efficient utilization of urban road capacity, effectively alleviating network-wide congestion while maintaining high throughput and travel efficiency.

%% file: sections/related_work.tex
\section{Related Work}
\subsection{Vehicle Navigation}
Early studies in vehicle routing primarily focused on pre-defined, static shortest-path search algorithms, where congestion effects were handled through dynamic traffic assignment \cite{peeta2001foundations}. Recent advances in learning-based routing move beyond static heuristics toward adaptive, congestion-aware control. For instance, deep reinforcement learning frameworks have achieved great success in improving connected vehicle navigation by dynamically adjusting routes for avoiding traffic congestion \cite{9714823}.
Building on these foundations, asynchronous multi-agent reinforcement learning (MARL) frameworks model city regions as independent yet interacting agents, enabling dynamic multi-source routing under complex urban flow conditions \cite{yin2025cooperative}. Large-scale MARL further demonstrates that even partial adoption of autonomous decision-making can significantly smooth congestion and enhance throughput \cite{cui2022scalablemultiagentdrivingpolicies}.
Meanwhile, hybrid RL–DTA approaches integrate system-optimal flow principles with data-driven learning to accelerate convergence and improve equilibrium stability \cite{wang2025reinforcementlearningbasedsequentialroute}. More recent hybrid control designs examine robustness and user compliance under mixed autonomy, underscoring reinforcement learning’s potential to generalize city-scale routing and coordination \cite{YUN2024104721}.

\subsection{Intelligent Transportation}
Scalable urban transportation control increasingly depends on hierarchical frameworks that align global coordination with local adaptability. Early centralized approaches focused on system-optimal routing and traffic signal optimization through dynamic traffic assignment and RL \cite{wang2025reinforcementlearningbasedsequentialroute,YUN2024104721}.
To extend scalability, hierarchical designs have emerged that couple macroscopic decision-making with microscopic control. For instance, NavTL \cite{10.1145/3580305.3599839} integrates city-level region guidance with vehicle and intersection control to mitigate congestion under mixed autonomy. Similarly, HiLight \cite{Xu_Wang_Wang_Jia_Lu_2021} employs a manager–worker hierarchy, where upper-level agents determine regional strategies and lower-level agents execute intersection-level control. These hierarchical MARL systems improve convergence stability, sample efficiency, and citywide travel time compared to flat or purely centralized schemes.
Such multi-layered coordination underscores the importance of balancing top-down optimization with bottom-up adaptability.

\subsection{Large Language Models for Decision Making}
Large language models (LLMs) have recently emerged as powerful cognitive agents capable of reasoning, planning, and coordination in dynamic environments \cite{jin2025comprehensivesurveymultiagentcooperative,chowa2025languageactionreviewlarge,tran2025multiagentcollaborationmechanismssurvey,bilal2025metathinkingllmsmultiagentreinforcement}.
Frameworks such as ReAct \cite{yao2023react} interleave reasoning and action to enhance interpretability and adaptability in sequential decision tasks \cite{yao2023react}. Building on this paradigm, MetaGPT encodes role-based standard operating procedures to enable collaborative task decomposition and cross-agent verification \cite{hong2024metagpt}. Meanwhile, generative agents simulate persistent memory, intent formation, and emergent social coordination within multi-agent societies \cite{10.1145/3586183.3606763}.
Together, these advances suggest that LLMs can evolve from reactive language processors into deliberative, world-model–driven agents—capable of integrating spatiotemporal reasoning and cooperative decision-making. This perspective directly informs our design of CityNav, where LLM-based agents serve as hierarchical planners that bridge individual navigation with system-level optimization.

%% file: sections/conclusion.tex
\section{Conclusion and Future Work}

This study presents CityNav, the first LLM-based hierarchical framework for large-scale multi-vehicle dynamic navigation. By coupling a global traffic allocation agent with decentralized local navigation agents, CityNav effectively bridges global coordination and local adaptability. The proposed cooperative reasoning optimization mechanism enables hierarchical agents to jointly reason, align local decisions with global objectives, and achieve a balance between individual vehicle efficiency and overall network mobility. Experimental results on multiple real-world urban networks demonstrate that CityNav significantly outperforms both classical path search and reinforcement learning approaches in travel efficiency, congestion reduction, and scalability. These findings underscore the potential of LLMs to enable adaptive, cooperative, and city-scale navigation through structured reasoning and communication.

As for future work, our framework primarily focuses on vehicle mobility, while it does not yet integrate multimodal traffic factors, such as pedestrians and public transportation. Future extensions will incorporate these heterogeneous urban contexts, enabling LLM agents to reason under broader traffic contexts and further advance intelligent urban mobility coordination.

%% file: sections/appendix.tex
\clearpage
\appendix
\section{Appendix}
\subsection{Prompt Template}
\begin{figure}[h]
    \centering
    \includegraphics[width=\linewidth]{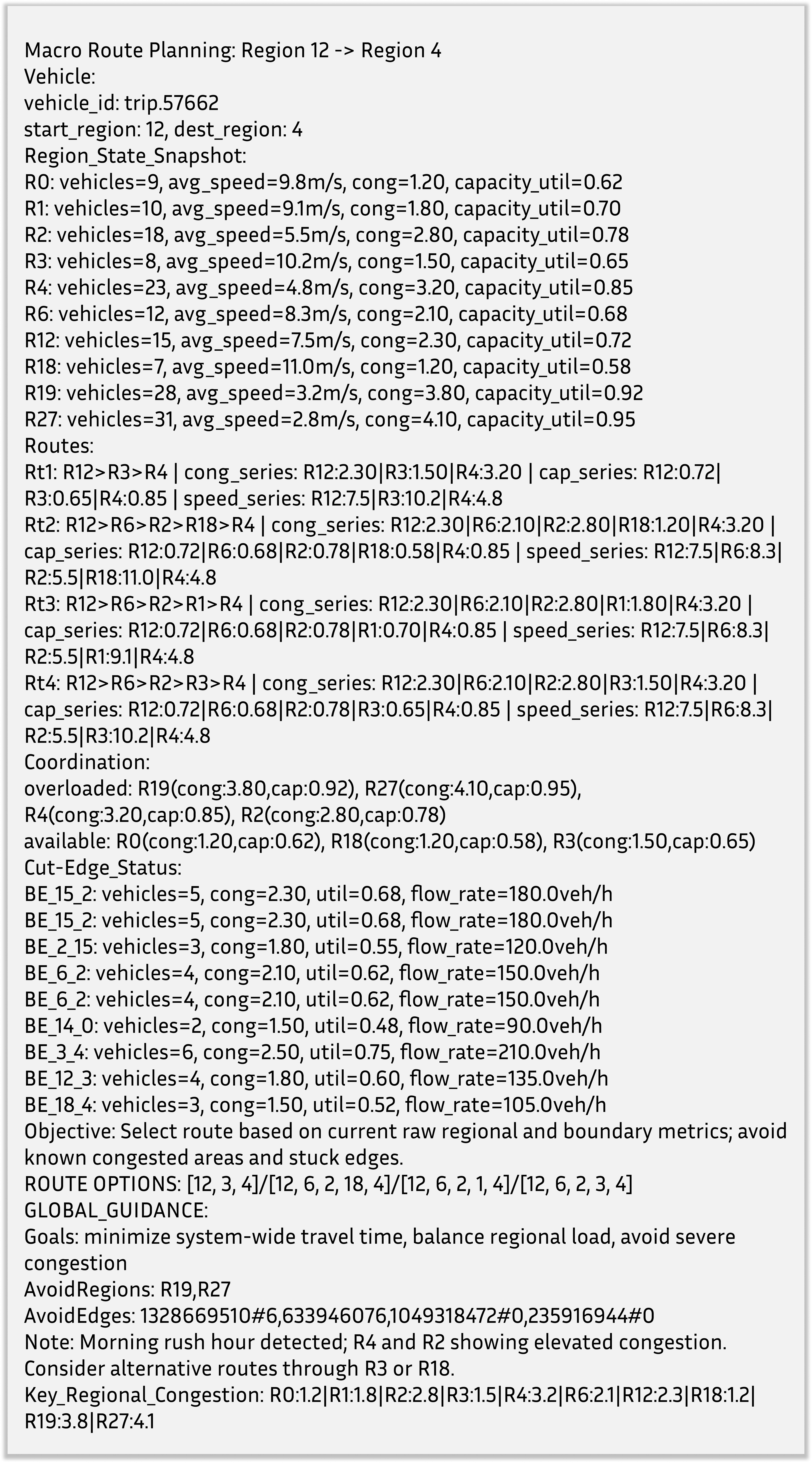}
    \caption{Global traffic allocation prompt.}
    \label{fig:Network-wide Traffic Allocation}
\end{figure}

\begin{figure}[h]
    \centering
    \includegraphics[width=\linewidth]{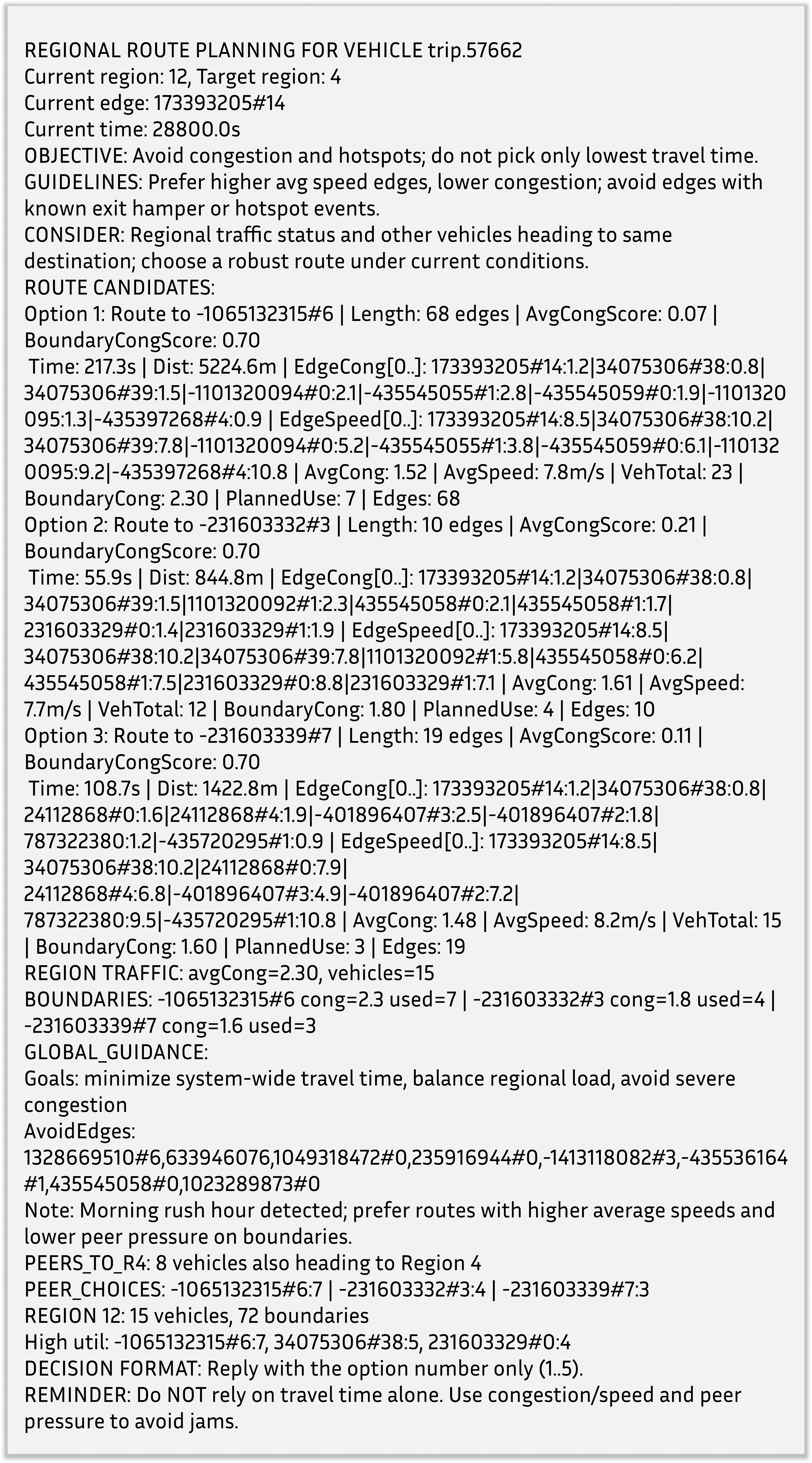}
    \caption{Local navigation prompt.}
    \label{fig:Globle Aware Intra-Region Navigation}
\end{figure}

We present the prompt template of global traffic allocation and local navigation agents in figure \ref{fig:Network-wide Traffic Allocation} and figure \ref{fig:Globle Aware Intra-Region Navigation}.

\newpage

%% file: main.bbl

\begin{thebibliography}{47}


\ifx \showCODEN    \undefined \def \showCODEN     #1{\unskip}     \fi
\ifx \showISBNx    \undefined \def \showISBNx     #1{\unskip}     \fi
\ifx \showISBNxiii \undefined \def \showISBNxiii  #1{\unskip}     \fi
\ifx \showISSN     \undefined \def \showISSN      #1{\unskip}     \fi
\ifx \showLCCN     \undefined \def \showLCCN      #1{\unskip}     \fi
\ifx \shownote     \undefined \def \shownote      #1{#1}          \fi
\ifx \showarticletitle \undefined \def \showarticletitle #1{#1}   \fi
\ifx \showURL      \undefined \def \showURL       {\relax}        \fi
\providecommand\bibfield[2]{#2}
\providecommand\bibinfo[2]{#2}
\providecommand\natexlab[1]{#1}
\providecommand\showeprint[2][]{arXiv:#2}

\bibitem[Ameli et~al\mbox{.}(2020)]%
        {https://doi.org/10.1111/mice.12577}
\bibfield{author}{\bibinfo{person}{Mostafa Ameli}, \bibinfo{person}{Jean-Patrick Lebacque}, {and} \bibinfo{person}{Ludovic Leclercq}.} \bibinfo{year}{2020}\natexlab{}.
\newblock \showarticletitle{Simulation-based dynamic traffic assignment: Meta-heuristic solution methods with parallel computing}.
\newblock \bibinfo{journal}{\emph{Computer-Aided Civil and Infrastructure Engineering}} \bibinfo{volume}{35}, \bibinfo{number}{10} (\bibinfo{year}{2020}), \bibinfo{pages}{1047--1062}.
\newblock


\bibitem[Anderson(2010)]%
        {NBERw16576}
\bibfield{author}{\bibinfo{person}{James~E Anderson}.} \bibinfo{year}{2010}\natexlab{}.
\newblock \bibinfo{booktitle}{\emph{The gravity model}}.
\newblock \bibinfo{type}{Working Paper} 16576. \bibinfo{institution}{National Bureau of Economic Research}.
\newblock


\bibitem[Arasteh et~al\mbox{.}(2022)]%
        {10.1145/3557915.3561005}
\bibfield{author}{\bibinfo{person}{Fazel Arasteh}, \bibinfo{person}{Soroush SheikhGarGar}, {and} \bibinfo{person}{Manos Papagelis}.} \bibinfo{year}{2022}\natexlab{}.
\newblock \showarticletitle{Network-aware multi-agent reinforcement learning for the vehicle navigation problem} \emph{(\bibinfo{series}{SIGSPATIAL '22})}. \bibinfo{publisher}{Association for Computing Machinery}, \bibinfo{address}{New York, NY, USA}, Article \bibinfo{articleno}{69}, \bibinfo{numpages}{4}~pages.
\newblock
\showISBNx{9781450395298}


\bibitem[Bilal et~al\mbox{.}(2025)]%
        {bilal2025metathinkingllmsmultiagentreinforcement}
\bibfield{author}{\bibinfo{person}{Ahsan Bilal}, \bibinfo{person}{Muhammad~Ahmed Mohsin}, \bibinfo{person}{Muhammad Umer}, \bibinfo{person}{Muhammad Awais~Khan Bangash}, {and} \bibinfo{person}{Muhammad~Ali Jamshed}.} \bibinfo{year}{2025}\natexlab{}.
\newblock \bibinfo{title}{Meta-thinking in LLMs via multi-agent reinforcement learning: A survey}.
\newblock
\showeprint[arxiv]{2504.14520}~[cs.AI]


\bibitem[Blondel et~al\mbox{.}(2008)]%
        {blondel2008fast}
\bibfield{author}{\bibinfo{person}{Vincent~D Blondel}, \bibinfo{person}{Jean-Loup Guillaume}, \bibinfo{person}{Renaud Lambiotte}, {and} \bibinfo{person}{Etienne Lefebvre}.} \bibinfo{year}{2008}\natexlab{}.
\newblock \showarticletitle{Fast unfolding of communities in large networks}.
\newblock \bibinfo{journal}{\emph{Journal of statistical mechanics: theory and experiment}} \bibinfo{volume}{2008}, \bibinfo{number}{10} (\bibinfo{year}{2008}), \bibinfo{pages}{P10008}.
\newblock


\bibitem[Box and Jenkins(1976)]%
        {box1976analysis}
\bibfield{author}{\bibinfo{person}{George Box} {and} \bibinfo{person}{GM Jenkins}.} \bibinfo{year}{1976}\natexlab{}.
\newblock \showarticletitle{Analysis: Forecasting and Control}.
\newblock \bibinfo{journal}{\emph{San francisco}} (\bibinfo{year}{1976}).
\newblock


\bibitem[Chowa et~al\mbox{.}(2025)]%
        {chowa2025languageactionreviewlarge}
\bibfield{author}{\bibinfo{person}{Sadia~Sultana Chowa}, \bibinfo{person}{Riasad Alvi}, \bibinfo{person}{Subhey~Sadi Rahman}, \bibinfo{person}{Md~Abdur Rahman}, \bibinfo{person}{Mohaimenul Azam~Khan Raiaan}, \bibinfo{person}{Md~Rafiqul Islam}, \bibinfo{person}{Mukhtar Hussain}, {and} \bibinfo{person}{Sami Azam}.} \bibinfo{year}{2025}\natexlab{}.
\newblock \bibinfo{title}{From language to action: A review of large language models as autonomous agents and tool users}.
\newblock
\showeprint[arxiv]{2508.17281}~[cs.CL]


\bibitem[{City of Chicago}(2025)]%
        {chicago_data_portal}
\bibfield{author}{\bibinfo{person}{{City of Chicago}}.} \bibinfo{year}{2025}\natexlab{}.
\newblock \bibinfo{title}{Chicago data portal}.
\newblock


\bibitem[{City of Chicago}(nd)]%
        {chicago_taxi_trips_2013_2023}
\bibfield{author}{\bibinfo{person}{{City of Chicago}}.} \bibinfo{year}{n.d.}\natexlab{}.
\newblock \bibinfo{title}{Taxi trips (2013--2023)}.
\newblock
\newblock
\shownote{Chicago Data Portal (dataset ID: wrvz-psew).}.


\bibitem[{City of New York}(2025)]%
        {nyc_opendata_portal}
\bibfield{author}{\bibinfo{person}{{City of New York}}.} \bibinfo{year}{2025}\natexlab{}.
\newblock \bibinfo{title}{NYC open data}.
\newblock


\bibitem[Cui et~al\mbox{.}(2022)]%
        {cui2022scalablemultiagentdrivingpolicies}
\bibfield{author}{\bibinfo{person}{Jiaxun Cui}, \bibinfo{person}{William Macke}, \bibinfo{person}{Harel Yedidsion}, \bibinfo{person}{Daniel Urieli}, {and} \bibinfo{person}{Peter Stone}.} \bibinfo{year}{2022}\natexlab{}.
\newblock \bibinfo{title}{Scalable multiagent driving policies for reducing traffic congestion}.
\newblock
\showeprint[arxiv]{2103.00058}~[cs.AI]


\bibitem[Dijkstra(2022)]%
        {dijkstra2022note}
\bibfield{author}{\bibinfo{person}{Edsger~W Dijkstra}.} \bibinfo{year}{2022}\natexlab{}.
\newblock \showarticletitle{A note on two problems in connexion with graphs}.
\newblock In \bibinfo{booktitle}{\emph{Edsger Wybe Dijkstra: his life, work, and legacy}}. \bibinfo{pages}{287--290}.
\newblock


\bibitem[Guo et~al\mbox{.}(2025)]%
        {guo2025deepseek}
\bibfield{author}{\bibinfo{person}{Daya Guo}, \bibinfo{person}{Dejian Yang}, \bibinfo{person}{Haowei Zhang}, \bibinfo{person}{Junxiao Song}, \bibinfo{person}{Ruoyu Zhang}, \bibinfo{person}{Runxin Xu}, \bibinfo{person}{Qihao Zhu}, \bibinfo{person}{Shirong Ma}, \bibinfo{person}{Peiyi Wang}, \bibinfo{person}{Xiao Bi}, {et~al\mbox{.}}} \bibinfo{year}{2025}\natexlab{}.
\newblock \showarticletitle{Deepseek-r1: Incentivizing reasoning capability in llms via reinforcement learning}.
\newblock \bibinfo{journal}{\emph{arXiv preprint arXiv:2501.12948}} (\bibinfo{year}{2025}).
\newblock


\bibitem[Hart et~al\mbox{.}(1968)]%
        {hart1968formal}
\bibfield{author}{\bibinfo{person}{Peter~E Hart}, \bibinfo{person}{Nils~J Nilsson}, {and} \bibinfo{person}{Bertram Raphael}.} \bibinfo{year}{1968}\natexlab{}.
\newblock \showarticletitle{A formal basis for the heuristic determination of minimum cost paths}.
\newblock \bibinfo{journal}{\emph{IEEE transactions on Systems Science and Cybernetics}} \bibinfo{volume}{4}, \bibinfo{number}{2} (\bibinfo{year}{1968}), \bibinfo{pages}{100--107}.
\newblock


\bibitem[Hong et~al\mbox{.}(2024)]%
        {hong2024metagpt}
\bibfield{author}{\bibinfo{person}{Sirui Hong}, \bibinfo{person}{Mingchen Zhuge}, \bibinfo{person}{Jonathan Chen}, \bibinfo{person}{Xiawu Zheng}, \bibinfo{person}{Yuheng Cheng}, \bibinfo{person}{Ceyao Zhang}, \bibinfo{person}{Jinlin Wang}, \bibinfo{person}{Zili Wang}, \bibinfo{person}{Steven Ka~Shing Yau}, \bibinfo{person}{Zijuan Lin}, {et~al\mbox{.}}} \bibinfo{year}{2024}\natexlab{}.
\newblock \showarticletitle{MetaGPT: Meta programming for a multi-agent collaborative framework}. International Conference on Learning Representations, ICLR.
\newblock


\bibitem[Huang et~al\mbox{.}(2025)]%
        {huang2025biomni}
\bibfield{author}{\bibinfo{person}{Kexin Huang}, \bibinfo{person}{Serena Zhang}, \bibinfo{person}{Hanchen Wang}, \bibinfo{person}{Yuanhao Qu}, \bibinfo{person}{Yingzhou Lu}, \bibinfo{person}{Yusuf Roohani}, \bibinfo{person}{Ryan Li}, \bibinfo{person}{Lin Qiu}, \bibinfo{person}{Junze Zhang}, \bibinfo{person}{Yin Di}, {et~al\mbox{.}}} \bibinfo{year}{2025}\natexlab{}.
\newblock \showarticletitle{Biomni: A general-purpose biomedical AI agent}.
\newblock \bibinfo{journal}{\emph{bioRxiv}} (\bibinfo{year}{2025}), \bibinfo{pages}{2025--05}.
\newblock


\bibitem[Ji et~al\mbox{.}(2020)]%
        {ji2020survey}
\bibfield{author}{\bibinfo{person}{Baofeng Ji}, \bibinfo{person}{Xueru Zhang}, \bibinfo{person}{Shahid Mumtaz}, \bibinfo{person}{Congzheng Han}, \bibinfo{person}{Chunguo Li}, \bibinfo{person}{Hong Wen}, {and} \bibinfo{person}{Dan Wang}.} \bibinfo{year}{2020}\natexlab{}.
\newblock \showarticletitle{Survey on the internet of vehicles: Network architectures and applications}.
\newblock \bibinfo{journal}{\emph{IEEE Communications Standards Magazine}} \bibinfo{volume}{4}, \bibinfo{number}{1} (\bibinfo{year}{2020}), \bibinfo{pages}{34--41}.
\newblock


\bibitem[Jin et~al\mbox{.}(2025)]%
        {jin2025comprehensivesurveymultiagentcooperative}
\bibfield{author}{\bibinfo{person}{Weiqiang Jin}, \bibinfo{person}{Hongyang Du}, \bibinfo{person}{Biao Zhao}, \bibinfo{person}{Xingwu Tian}, \bibinfo{person}{Bohang Shi}, {and} \bibinfo{person}{Guang Yang}.} \bibinfo{year}{2025}\natexlab{}.
\newblock \bibinfo{title}{A Comprehensive Survey on Multi-Agent Cooperative Decision-Making: Scenarios, Approaches, Challenges and Perspectives}.
\newblock
\showeprint[arxiv]{2503.13415}~[cs.MA]


\bibitem[Koh et~al\mbox{.}(2020)]%
        {koh2020real}
\bibfield{author}{\bibinfo{person}{Songsang Koh}, \bibinfo{person}{Bo Zhou}, \bibinfo{person}{Hui Fang}, \bibinfo{person}{Po Yang}, \bibinfo{person}{Zaili Yang}, \bibinfo{person}{Qiang Yang}, \bibinfo{person}{Lin Guan}, {and} \bibinfo{person}{Zhigang Ji}.} \bibinfo{year}{2020}\natexlab{}.
\newblock \showarticletitle{Real-time deep reinforcement learning based vehicle navigation}.
\newblock \bibinfo{journal}{\emph{Applied Soft Computing}}  \bibinfo{volume}{96} (\bibinfo{year}{2020}), \bibinfo{pages}{106694}.
\newblock


\bibitem[Lai et~al\mbox{.}(2024)]%
        {lai2024llmlightlargelanguagemodels}
\bibfield{author}{\bibinfo{person}{Siqi Lai}, \bibinfo{person}{Zhao Xu}, \bibinfo{person}{Weijia Zhang}, \bibinfo{person}{Hao Liu}, {and} \bibinfo{person}{Hui Xiong}.} \bibinfo{year}{2024}\natexlab{}.
\newblock \bibinfo{title}{LLMLight: Large language models as traffic signal control agents}.
\newblock
\showeprint[arxiv]{2312.16044}~[cs.AI]


\bibitem[Leclercq et~al\mbox{.}(2021)]%
        {LECLERCQ2021118}
\bibfield{author}{\bibinfo{person}{Ludovic Leclercq}, \bibinfo{person}{Andres Ladino}, {and} \bibinfo{person}{Cécile Becarie}.} \bibinfo{year}{2021}\natexlab{}.
\newblock \showarticletitle{Enforcing optimal routing through dynamic avoidance maps}.
\newblock \bibinfo{journal}{\emph{Transportation Research Part B: Methodological}}  \bibinfo{volume}{149} (\bibinfo{year}{2021}), \bibinfo{pages}{118--137}.
\newblock
\showISSN{0191-2615}


\bibitem[Li et~al\mbox{.}(2021)]%
        {ijcai2021p503}
\bibfield{author}{\bibinfo{person}{Ke Li}, \bibinfo{person}{Lisi Chen}, \bibinfo{person}{Shuo Shang}, \bibinfo{person}{Panos Kalnis}, {and} \bibinfo{person}{Bin Yao}.} \bibinfo{year}{2021}\natexlab{}.
\newblock \showarticletitle{Traffic congestion alleviation over dynamic road networks: continuous optimal route combination for trip query streams}. In \bibinfo{booktitle}{\emph{Proceedings of the Thirtieth International Joint Conference on Artificial Intelligence, {IJCAI-21}}}, \bibfield{editor}{\bibinfo{person}{Zhi-Hua Zhou}} (Ed.). \bibinfo{publisher}{International Joint Conferences on Artificial Intelligence Organization}, \bibinfo{pages}{3656--3662}.
\newblock
\newblock
\shownote{Main Track}.


\bibitem[Liu et~al\mbox{.}(2025)]%
        {liu2025mm}
\bibfield{author}{\bibinfo{person}{Fan Liu}, \bibinfo{person}{Zherui Yang}, \bibinfo{person}{Cancheng Liu}, \bibinfo{person}{Tianrui Song}, \bibinfo{person}{Xiaofeng Gao}, {and} \bibinfo{person}{Hao Liu}.} \bibinfo{year}{2025}\natexlab{}.
\newblock \showarticletitle{Mm-agent: Llm as agents for real-world mathematical modeling problem}.
\newblock \bibinfo{journal}{\emph{arXiv preprint arXiv:2505.14148}} (\bibinfo{year}{2025}).
\newblock


\bibitem[Lopez et~al\mbox{.}(2018)]%
        {SUMO2018}
\bibfield{author}{\bibinfo{person}{Pablo~Alvarez Lopez}, \bibinfo{person}{Michael Behrisch}, \bibinfo{person}{Laura Bieker-Walz}, \bibinfo{person}{Jakob Erdmann}, \bibinfo{person}{Yun-Pang Fl{\"o}tter{\"o}d}, \bibinfo{person}{Robert Hilbrich}, \bibinfo{person}{Leonhard L{\"u}cken}, \bibinfo{person}{Johannes Rummel}, \bibinfo{person}{Peter Wagner}, {and} \bibinfo{person}{Evamarie Wie{\ss}ner}.} \bibinfo{year}{2018}\natexlab{}.
\newblock \showarticletitle{Microscopic traffic simulation using SUMO}, In \bibinfo{booktitle}{The 21st IEEE International Conference on Intelligent Transportation Systems}.
\newblock \bibinfo{journal}{\emph{IEEE Intelligent Transportation Systems Conference (ITSC)}}.
\newblock


\bibitem[Luo et~al\mbox{.}(2023)]%
        {Luo_Wang_Zhang_Yuan_Li_2023}
\bibfield{author}{\bibinfo{person}{Guiyang Luo}, \bibinfo{person}{Yantao Wang}, \bibinfo{person}{Hui Zhang}, \bibinfo{person}{Quan Yuan}, {and} \bibinfo{person}{Jinglin Li}.} \bibinfo{year}{2023}\natexlab{}.
\newblock \showarticletitle{AlphaRoute: Large-scale coordinated route planning via monte carlo tree search}.
\newblock \bibinfo{journal}{\emph{Proceedings of the AAAI Conference on Artificial Intelligence}} \bibinfo{volume}{37}, \bibinfo{number}{10} (\bibinfo{date}{Jun.} \bibinfo{year}{2023}), \bibinfo{pages}{12058--12067}.
\newblock


\bibitem[{New York City Department of Transportation}(nd)]%
        {nyc_citywide_mobility_survey}
\bibfield{author}{\bibinfo{person}{{New York City Department of Transportation}}.} \bibinfo{year}{n.d.}\natexlab{}.
\newblock \bibinfo{title}{Citywide mobility survey}.
\newblock


\bibitem[{New york city taxi and limousine commission (TLC)}(2016)]%
        {nyc_tlc_green_taxi_2015}
\bibfield{author}{\bibinfo{person}{{New york city taxi and limousine commission (TLC)}}.} \bibinfo{year}{2016}\natexlab{}.
\newblock \bibinfo{title}{2015 Green taxi trip data}.
\newblock
\newblock
\shownote{NYC Open Data (dataset ID: gi8d-wdg5).}.


\bibitem[Park et~al\mbox{.}(2023)]%
        {10.1145/3586183.3606763}
\bibfield{author}{\bibinfo{person}{Joon~Sung Park}, \bibinfo{person}{Joseph O'Brien}, \bibinfo{person}{Carrie~Jun Cai}, \bibinfo{person}{Meredith~Ringel Morris}, \bibinfo{person}{Percy Liang}, {and} \bibinfo{person}{Michael~S. Bernstein}.} \bibinfo{year}{2023}\natexlab{}.
\newblock \showarticletitle{Generative agents: interactive simulacra of human behavior}. In \bibinfo{booktitle}{\emph{Proceedings of the 36th Annual ACM Symposium on User Interface Software and Technology}} (San Francisco, CA, USA) \emph{(\bibinfo{series}{UIST '23})}. \bibinfo{publisher}{Association for Computing Machinery}, \bibinfo{address}{New York, NY, USA}, Article \bibinfo{articleno}{2}, \bibinfo{numpages}{22}~pages.
\newblock
\showISBNx{9798400701320}


\bibitem[Peeta and Ziliaskopoulos(2001)]%
        {peeta2001foundations}
\bibfield{author}{\bibinfo{person}{Srinivas Peeta} {and} \bibinfo{person}{Athanasios~K Ziliaskopoulos}.} \bibinfo{year}{2001}\natexlab{}.
\newblock \showarticletitle{Foundations of dynamic traffic assignment: The past, the present and the future}.
\newblock \bibinfo{journal}{\emph{Networks and spatial economics}} \bibinfo{volume}{1}, \bibinfo{number}{3} (\bibinfo{year}{2001}), \bibinfo{pages}{233--265}.
\newblock


\bibitem[Ren et~al\mbox{.}(2022)]%
        {9714823}
\bibfield{author}{\bibinfo{person}{Lei Ren}, \bibinfo{person}{Xiaoyang Fan}, \bibinfo{person}{Jin Cui}, \bibinfo{person}{Zhen Shen}, \bibinfo{person}{Yisheng Lv}, {and} \bibinfo{person}{Gang Xiong}.} \bibinfo{year}{2022}\natexlab{}.
\newblock \showarticletitle{A multi-agent reinforcement learning method with route recorders for vehicle routing in supply chain management}.
\newblock \bibinfo{journal}{\emph{IEEE Transactions on Intelligent Transportation Systems}} \bibinfo{volume}{23}, \bibinfo{number}{9} (\bibinfo{year}{2022}), \bibinfo{pages}{16410--16420}.
\newblock


\bibitem[Su et~al\mbox{.}(2024)]%
        {su2024many}
\bibfield{author}{\bibinfo{person}{Haoyang Su}, \bibinfo{person}{Renqi Chen}, \bibinfo{person}{Shixiang Tang}, \bibinfo{person}{Zhenfei Yin}, \bibinfo{person}{Xinzhe Zheng}, \bibinfo{person}{Jinzhe Li}, \bibinfo{person}{Biqing Qi}, \bibinfo{person}{Qi Wu}, \bibinfo{person}{Hui Li}, \bibinfo{person}{Wanli Ouyang}, {et~al\mbox{.}}} \bibinfo{year}{2024}\natexlab{}.
\newblock \showarticletitle{Many heads are better than one: Improved scientific idea generation by A LLM-Based multi-agent system}.
\newblock \bibinfo{journal}{\emph{arXiv preprint arXiv:2410.09403}} (\bibinfo{year}{2024}).
\newblock


\bibitem[Sun et~al\mbox{.}(2023a)]%
        {sun2023hierarchical}
\bibfield{author}{\bibinfo{person}{Qian Sun}, \bibinfo{person}{Le Zhang}, \bibinfo{person}{Huan Yu}, \bibinfo{person}{Weijia Zhang}, \bibinfo{person}{Yu Mei}, {and} \bibinfo{person}{Hui Xiong}.} \bibinfo{year}{2023}\natexlab{a}.
\newblock \showarticletitle{Hierarchical reinforcement learning for dynamic autonomous vehicle navigation at intelligent intersections}. In \bibinfo{booktitle}{\emph{Proceedings of the 29th ACM SIGKDD Conference on Knowledge Discovery and Data Mining}}. \bibinfo{pages}{4852--4861}.
\newblock


\bibitem[Sun et~al\mbox{.}(2023b)]%
        {10.1145/3580305.3599839}
\bibfield{author}{\bibinfo{person}{Qian Sun}, \bibinfo{person}{Le Zhang}, \bibinfo{person}{Huan Yu}, \bibinfo{person}{Weijia Zhang}, \bibinfo{person}{Yu Mei}, {and} \bibinfo{person}{Hui Xiong}.} \bibinfo{year}{2023}\natexlab{b}.
\newblock \showarticletitle{Hierarchical reinforcement learning for dynamic autonomous vehicle navigation at intelligent intersections}. In \bibinfo{booktitle}{\emph{Proceedings of the 29th ACM SIGKDD Conference on Knowledge Discovery and Data Mining}} (Long Beach, CA, USA) \emph{(\bibinfo{series}{KDD '23})}. \bibinfo{publisher}{Association for Computing Machinery}, \bibinfo{address}{New York, NY, USA}, \bibinfo{pages}{4852–4861}.
\newblock
\showISBNx{9798400701030}


\bibitem[Tran et~al\mbox{.}(2025)]%
        {tran2025multiagentcollaborationmechanismssurvey}
\bibfield{author}{\bibinfo{person}{Khanh-Tung Tran}, \bibinfo{person}{Dung Dao}, \bibinfo{person}{Minh-Duong Nguyen}, \bibinfo{person}{Quoc-Viet Pham}, \bibinfo{person}{Barry O'Sullivan}, {and} \bibinfo{person}{Hoang~D. Nguyen}.} \bibinfo{year}{2025}\natexlab{}.
\newblock \bibinfo{title}{Multi-agent collaboration mechanisms: A survey of LLMs}.
\newblock
\showeprint[arxiv]{2501.06322}~[cs.AI]


\bibitem[Wan et~al\mbox{.}(2025)]%
        {wan2025rema}
\bibfield{author}{\bibinfo{person}{Ziyu Wan}, \bibinfo{person}{Yunxiang Li}, \bibinfo{person}{Xiaoyu Wen}, \bibinfo{person}{Yan Song}, \bibinfo{person}{Hanjing Wang}, \bibinfo{person}{Linyi Yang}, \bibinfo{person}{Mark Schmidt}, \bibinfo{person}{Jun Wang}, \bibinfo{person}{Weinan Zhang}, \bibinfo{person}{Shuyue Hu}, {et~al\mbox{.}}} \bibinfo{year}{2025}\natexlab{}.
\newblock \showarticletitle{Rema: Learning to meta-think for llms with multi-agent reinforcement learning}.
\newblock \bibinfo{journal}{\emph{arXiv preprint arXiv:2503.09501}} (\bibinfo{year}{2025}).
\newblock


\bibitem[Wang et~al\mbox{.}(2019)]%
        {wang2019survey}
\bibfield{author}{\bibinfo{person}{Jian Wang}, \bibinfo{person}{Yameng Shao}, \bibinfo{person}{Yuming Ge}, {and} \bibinfo{person}{Rundong Yu}.} \bibinfo{year}{2019}\natexlab{}.
\newblock \showarticletitle{A survey of vehicle to everything (V2X) testing}.
\newblock \bibinfo{journal}{\emph{Sensors}} \bibinfo{volume}{19}, \bibinfo{number}{2} (\bibinfo{year}{2019}), \bibinfo{pages}{334}.
\newblock


\bibitem[Wang et~al\mbox{.}(2025)]%
        {wang2025reinforcementlearningbasedsequentialroute}
\bibfield{author}{\bibinfo{person}{Leizhen Wang}, \bibinfo{person}{Peibo Duan}, \bibinfo{person}{Cheng Lyu}, {and} \bibinfo{person}{Zhenliang Ma}.} \bibinfo{year}{2025}\natexlab{}.
\newblock \bibinfo{title}{Reinforcement learning-based sequential route recommendation for system-optimal traffic assignment}.
\newblock
\showeprint[arxiv]{2505.20889}~[cs.AI]


\bibitem[Wang et~al\mbox{.}(2022)]%
        {WANG2022103444}
\bibfield{author}{\bibinfo{person}{Yibing Wang}, \bibinfo{person}{Mingming Zhao}, \bibinfo{person}{Xianghua Yu}, \bibinfo{person}{Yonghui Hu}, \bibinfo{person}{Pengjun Zheng}, \bibinfo{person}{Wei Hua}, \bibinfo{person}{Lihui Zhang}, \bibinfo{person}{Simon Hu}, {and} \bibinfo{person}{Jingqiu Guo}.} \bibinfo{year}{2022}\natexlab{}.
\newblock \showarticletitle{Real-time joint traffic state and model parameter estimation on freeways with fixed sensors and connected vehicles: State-of-the-art overview, methods, and case studies}.
\newblock \bibinfo{journal}{\emph{Transportation Research Part C: Emerging Technologies}}  \bibinfo{volume}{134} (\bibinfo{year}{2022}), \bibinfo{pages}{103444}.
\newblock
\showISSN{0968-090X}


\bibitem[Wang and Wang(2022)]%
        {wang2022xrouting}
\bibfield{author}{\bibinfo{person}{Zheng Wang} {and} \bibinfo{person}{Shen Wang}.} \bibinfo{year}{2022}\natexlab{}.
\newblock \showarticletitle{Xrouting: Explainable vehicle rerouting for urban road congestion avoidance using deep reinforcement learning}. In \bibinfo{booktitle}{\emph{2022 IEEE International Smart Cities Conference (ISC2)}}. IEEE, \bibinfo{pages}{1--7}.
\newblock


\bibitem[Xu et~al\mbox{.}(2021)]%
        {Xu_Wang_Wang_Jia_Lu_2021}
\bibfield{author}{\bibinfo{person}{Bingyu Xu}, \bibinfo{person}{Yaowei Wang}, \bibinfo{person}{Zhaozhi Wang}, \bibinfo{person}{Huizhu Jia}, {and} \bibinfo{person}{Zongqing Lu}.} \bibinfo{year}{2021}\natexlab{}.
\newblock \showarticletitle{Hierarchically and cooperatively learning traffic signal control}.
\newblock   \bibinfo{volume}{35} (\bibinfo{date}{May} \bibinfo{year}{2021}), \bibinfo{pages}{669--677}.
\newblock


\bibitem[Yang et~al\mbox{.}(2025)]%
        {yang2025qwen3technicalreport}
\bibfield{author}{\bibinfo{person}{An Yang}, \bibinfo{person}{Anfeng Li}, \bibinfo{person}{Baosong Yang}, \bibinfo{person}{Beichen Zhang}, \bibinfo{person}{Binyuan Hui}, \bibinfo{person}{Bo Zheng}, \bibinfo{person}{Bowen Yu}, \bibinfo{person}{Chang Gao}, \bibinfo{person}{Chengen Huang}, \bibinfo{person}{Chenxu Lv}, \bibinfo{person}{Chujie Zheng}, \bibinfo{person}{Dayiheng Liu}, \bibinfo{person}{Fan Zhou}, \bibinfo{person}{Fei Huang}, \bibinfo{person}{Feng Hu}, \bibinfo{person}{Hao Ge}, \bibinfo{person}{Haoran Wei}, \bibinfo{person}{Huan Lin}, \bibinfo{person}{Jialong Tang}, \bibinfo{person}{Jian Yang}, \bibinfo{person}{Jianhong Tu}, \bibinfo{person}{Jianwei Zhang}, \bibinfo{person}{Jianxin Yang}, \bibinfo{person}{Jiaxi Yang}, \bibinfo{person}{Jing Zhou}, \bibinfo{person}{Jingren Zhou}, \bibinfo{person}{Junyang Lin}, \bibinfo{person}{Kai Dang}, \bibinfo{person}{Keqin Bao}, \bibinfo{person}{Kexin Yang}, \bibinfo{person}{Le Yu}, \bibinfo{person}{Lianghao Deng}, \bibinfo{person}{Mei Li}, \bibinfo{person}{Mingfeng
  Xue}, \bibinfo{person}{Mingze Li}, \bibinfo{person}{Pei Zhang}, \bibinfo{person}{Peng Wang}, \bibinfo{person}{Qin Zhu}, \bibinfo{person}{Rui Men}, \bibinfo{person}{Ruize Gao}, \bibinfo{person}{Shixuan Liu}, \bibinfo{person}{Shuang Luo}, \bibinfo{person}{Tianhao Li}, \bibinfo{person}{Tianyi Tang}, \bibinfo{person}{Wenbiao Yin}, \bibinfo{person}{Xingzhang Ren}, \bibinfo{person}{Xinyu Wang}, \bibinfo{person}{Xinyu Zhang}, \bibinfo{person}{Xuancheng Ren}, \bibinfo{person}{Yang Fan}, \bibinfo{person}{Yang Su}, \bibinfo{person}{Yichang Zhang}, \bibinfo{person}{Yinger Zhang}, \bibinfo{person}{Yu Wan}, \bibinfo{person}{Yuqiong Liu}, \bibinfo{person}{Zekun Wang}, \bibinfo{person}{Zeyu Cui}, \bibinfo{person}{Zhenru Zhang}, \bibinfo{person}{Zhipeng Zhou}, {and} \bibinfo{person}{Zihan Qiu}.} \bibinfo{year}{2025}\natexlab{}.
\newblock \bibinfo{title}{Qwen3 technical report}.
\newblock
\showeprint[arxiv]{2505.09388}~[cs.CL]


\bibitem[Yao et~al\mbox{.}(2025)]%
        {yao2025comal}
\bibfield{author}{\bibinfo{person}{Huaiyuan Yao}, \bibinfo{person}{Longchao Da}, \bibinfo{person}{Vishnu Nandam}, \bibinfo{person}{Justin Turnau}, \bibinfo{person}{Zhiwei Liu}, \bibinfo{person}{Linsey Pang}, {and} \bibinfo{person}{Hua Wei}.} \bibinfo{year}{2025}\natexlab{}.
\newblock \showarticletitle{Comal: Collaborative multi-agent large language models for mixed-autonomy traffic}. In \bibinfo{booktitle}{\emph{Proceedings of the 2025 SIAM International Conference on Data Mining (SDM)}}. SIAM, \bibinfo{pages}{409--418}.
\newblock


\bibitem[Yao et~al\mbox{.}(2023)]%
        {yao2023react}
\bibfield{author}{\bibinfo{person}{Shunyu Yao}, \bibinfo{person}{Jeffrey Zhao}, \bibinfo{person}{Dian Yu}, \bibinfo{person}{Nan Du}, \bibinfo{person}{Izhak Shafran}, \bibinfo{person}{Karthik Narasimhan}, {and} \bibinfo{person}{Yuan Cao}.} \bibinfo{year}{2023}\natexlab{}.
\newblock \showarticletitle{React: Synergizing reasoning and acting in language models}. In \bibinfo{booktitle}{\emph{International Conference on Learning Representations (ICLR)}}.
\newblock


\bibitem[Yin et~al\mbox{.}(2025)]%
        {yin2025cooperative}
\bibfield{author}{\bibinfo{person}{Jiaming Yin}, \bibinfo{person}{Weixiong Rao}, \bibinfo{person}{Yu Xiao}, {and} \bibinfo{person}{Keshuang Tang}.} \bibinfo{year}{2025}\natexlab{}.
\newblock \showarticletitle{Cooperative path planning with asynchronous multiagent reinforcement learning}.
\newblock \bibinfo{journal}{\emph{IEEE Transactions on Mobile Computing}} (\bibinfo{year}{2025}).
\newblock


\bibitem[Yuan et~al\mbox{.}(2025)]%
        {yuan2025collmlight}
\bibfield{author}{\bibinfo{person}{Zirui Yuan}, \bibinfo{person}{Siqi Lai}, {and} \bibinfo{person}{Hao Liu}.} \bibinfo{year}{2025}\natexlab{}.
\newblock \showarticletitle{Collmlight: Cooperative large language model agents for network-wide traffic signal control}.
\newblock \bibinfo{journal}{\emph{arXiv preprint arXiv:2503.11739}} (\bibinfo{year}{2025}).
\newblock


\bibitem[Yun et~al\mbox{.}(2024)]%
        {YUN2024104721}
\bibfield{author}{\bibinfo{person}{Hyunsoo Yun}, \bibinfo{person}{Eui jin Kim}, \bibinfo{person}{Seung~Woo Ham}, {and} \bibinfo{person}{Dong-Kyu Kim}.} \bibinfo{year}{2024}\natexlab{}.
\newblock \showarticletitle{Navigating the non-compliance effects on system optimal route guidance using reinforcement learning}.
\newblock \bibinfo{journal}{\emph{Transportation Research Part C: Emerging Technologies}}  \bibinfo{volume}{165} (\bibinfo{year}{2024}), \bibinfo{pages}{104721}.
\newblock
\showISSN{0968-090X}


\bibitem[Zhao et~al\mbox{.}(2024)]%
        {zhao2024origin}
\bibfield{author}{\bibinfo{person}{Dong Zhao}, \bibinfo{person}{Adriana-Simona Mih{\u{a}}i{\c{t}}{\u{a}}}, \bibinfo{person}{Yuming Ou}, \bibinfo{person}{Hanna Grzybowska}, {and} \bibinfo{person}{Mo Li}.} \bibinfo{year}{2024}\natexlab{}.
\newblock \showarticletitle{Origin--destination matrix estimation for public transport: A multi-modal weighted graph approach}.
\newblock \bibinfo{journal}{\emph{Transportation Research Part C: Emerging Technologies}}  \bibinfo{volume}{165} (\bibinfo{year}{2024}), \bibinfo{pages}{104694}.
\newblock


\end{thebibliography}
